\documentclass[pdflatex,sn-nature]{sn-jnl}

\usepackage[utf8]{inputenc} 
\usepackage[T1]{fontenc}    
\usepackage{textcomp}       

\usepackage{amsmath}
\usepackage{amssymb}
\usepackage{amsfonts}
\usepackage{amsthm}
\usepackage{mathrsfs}

\usepackage{graphicx}
\usepackage{xcolor}
\usepackage{colortbl} 
\usepackage{bookmark}
\usepackage{placeins}

\usepackage{booktabs}
\usepackage{tabularx}
\usepackage{array}
\usepackage{multirow}
\usepackage{siunitx}

\usepackage{algorithm}
\usepackage{algpseudocode} 
\usepackage{listings}

\usepackage[title]{appendix}
\usepackage{manyfoot}

\usepackage{pifont}

\newcommand{\cmark}{\ding{51}}
\newcommand{\xmark}{\ding{55}}
\newcommand{\pmark}{\ensuremath{\sim}}

\usepackage{todonotes}

\theoremstyle{thmstyleone}%
\theoremstyle{thmstyletwo}%

\theoremstyle{thmstylethree}%

\begin{document}

\title[Article Title]{ElectroChemLLM: A Multimodal LLM Pipeline for High-Fidelity Electrochemical Informatics}


\author[2]{\fnm{Dikshya} \sur{Mohanty}}\email{dimohanty@cs.stonybrook.edu} \equalcont{These authors contributed equally to this work.}
\author[2]{\fnm{Manoj Praveen} \sur{Nandigama}}\email{manojpraveen.nandigama@stonybrook.edu} \equalcont{These authors contributed equally to this work.}
\author[1]{\fnm{Kuldeepsinh} \sur{Raj}}
\author[1]{\fnm{Sourav Kanti} \sur{Maiti}}

\author[2]{\fnm{Jiannan} \sur{Yang}}
\author[2]{\fnm{Ruiqi} \sur{Luo}}
\author[2]{\fnm{Srinivasa Phani Madhav} \sur{Marupudi}}
\author[2]{\fnm{Mounika} \sur{Geriki}}
\author[4]{\fnm{Prayag} \sur{Raghunandan}}
\author[2]{\fnm{Puneet} \sur{Lingutla}}

\author[3]{\fnm{Tengfei} \sur{Ma}}
\author[2]{\fnm{Niranjan} \sur{Balasubramanian}}

\author*[1]{\fnm{Nav Nidhi} \sur{Rajput}}\email{navnidhi.rajput@stonybrook.edu}

\affil*[1]{\orgdiv{Dept. of Materials Science and Chemical Engineering},
  \orgname{Stony Brook University}}

\affil[2]{\orgdiv{Dept. of Computer Science},
  \orgname{Stony Brook University}}

\affil[3]{\orgdiv{Dept. of Biomedical Informatics},
  \orgname{Stony Brook University}}

\affil[4]{\orgdiv{Dept. of Materials Engineering},
  \orgname{Indian Institute of Technology Jodhpur}}

\abstract{Electrochemistry is central to solving many modern-day challenges, including energy storage systems for a sustainable future, and its advancement relies heavily on the data from past findings. Existing automated pipelines are limited by text-only architectures, context fragmentation, and numerical hallucinations - collectively preventing accurate, large-scale curation of property databases. We introduce \texttt{ElectrochemLLM}, a multimodal information extraction pipeline that jointly identifies structured electrochemical records encompassing electrolyte compositions, electrode materials, cell configurations, and their measured physicochemical properties from unstructured scientific PDFs at scale. By combining hybrid document parsing with vision-capable LLM orchestration, the pipeline extracts verifiable, structured data across text, tables, and figures. Evaluated on the battery-electrolyte subset of our manually annotated benchmark of 232 papers, \texttt{ElectrochemLLM} achieves an F-1 score of 0.97, and has been applied to curate ElectroChemDB, a structured database spanning 15,000 electrochemical research articles. We demonstrate the utility of this resource by training machine learning models for electrolyte property prediction, illustrating a complete pathway from unstructured literature to predictive materials discovery.}


\vspace{-2em}

\maketitle

\section{Introduction}
\label{introduction}



Electrochemical technologies (next-generation battery electrolytes and $\text{CO}_2$ electro-reduction catalysts) are fundamental to the global transition towards long-term sustainable and carbon-neutral storage and chemical synthesis solutions. Designing such high-performance electrolytes, whether for lithium-ion, solid-state, or flow battery systems, requires precise and extensive knowledge of physicochemical properties such as ionic conductivity, electrochemical stability window, viscosity, and Coulombic efficiency, and how to achieve them. Similarly for $\text{CO}_2$ electro-reduction, advances rely on identifying electrolyte compositions that maximize Faradaic efficiency and product selectivity under feasible conditions~\cite{raj2026co2er}. In both cases, the design process begins with prior work. The relevant experimental data and corresponding conditions have been reported, but they remains largely siloed and often inaccessible, buried in unstructured text, tables, and figures from a scientific literature base that now exceeds five million articles and continues to grow at an exponential rate~\cite{historical_data_mining}. 

Manually curating such a corpus is impractical -- while expert annotation achieves high fidelity, given the scale of the literature, this is expensive and time consuming. Existing curated databases cover only a narrow fraction of published experimental data. Automated text-mining is an established alternative~\cite{Xiang}. ChemDataExtractor~\cite{ChemDataExtractor} provides extensible NLP pipelines for extracting chemical entities and properties from text, but requires task-specific models for new domains, and BatteryDataExtractor~\cite{huang2022batterydataextractor} extends text-based extraction to battery materials by combining BatteryBERT~\cite{batterybert} models with question-answering methods to recover material--property relationships. Encoders pre-trained on scientific text, such as SciBERT~\cite{Beltagy2019SciBERT},
MatBERT~\cite{MatBERT}, and MaterialsBERT~\cite{MaterialsBERT}, improve named entity recognition and relation classification over general-purpose models~\cite{MatSciBERT,MatSciNLP2023}. Domain-adapted large language models (LLMs) such as LLaMat~\cite{mishra2025foundationallargelanguagemodels} and OmniScience~\cite{prabhakar2025omnisciencedomainspecializedllmscientific} have extended this approach to broader range of materials science tasks such as structured extraction and scientific reasoning. However, these architectures are inherently text-only and share a critical limitation -- the quantitative property values most relevant for database curation are often reported only in figures, or in tables whose structures is lost on conversion to text, and so they lie largely beyond the reach of test-based models.

Multimodal LLMs have begun to close this gap, and are being applied beyond text to heterogeneous scientific imagery: Li and Jain~\cite{Lijain2026} diagnose visible, electroluminescence and infrared photovoltaic images with a single prompt-driven pipeline and no modality-specific retraining, returning structured outputs with confidence and rationale. While their task is classification, the capabilities of LLMs are now being extended to read figures and tables. Moreira-Filho et al.~\cite{Moreira-Filho} pair document-parsing tools with LLMs in a low-code KNIME workflow with separate text and image extraction modes, but the workflow is domain-agnostic, requiring the mode to be chosen per document, and outputs structured JSON without the chemistry-aware standardization required for a machine-learning-ready property database. MERMaid~\cite{MERMaid} applies vision-language models to figures, tables and captions to build knowledge graphs and has so far been demonstrated only on reaction mining. 

In the battery domain, LLMB~\cite{lee2026llmb} combines hierarchical LLM-based text mining with graph to extract lithium-metal-battery compositions and cycling data, but is limited to this specific battery chemistry and focuses on complete cells and cycling curves rather than formulation-level electrolyte properties. Table~\ref{tab:electrochemllm_comparison} summarises the capabilities and limitations of these approaches relative to \texttt{ElectroChemLLM}. None of the existing systems combine multimodal extraction with chemistry-aware standardisation at the level of electrolyte formulations and their measured properties.

Another issue with existing extraction pipelines is context fragmentation i.e. PDF text is processed at a sentence- or paragraph-level processing to manage context window limits. This breaks the hierarchical associations inherent to scientific documents, where an electrolyte introduced in one section may only acquire its measured property value in a results table several pages later. ChemDataExtractor and BatteryDataExtractor process isolated text spans and cannot link an entity to properties introduced elsewhere in the document. More recent LLM-based frameworks reason over larger bodies of knowledge. JRLKG~\cite{lei2024material} builds a knowlegde graph from the battery literature and prompts GPT-4 to reason jointly over it when predicting material properties, and (LLaMP)~\cite{chiang2024llamp} uses hierarchical agents to answer questions against materials projects. Both are built for prediction and question answering, rather than data curation. Further these generative architectures introduce a distinct failure mode: numerical hallucinations, where the model produces plausible but fabricated quantitative values. For large-scale, reliable curation of a composition and property database, explicit grounding mechanisms therefore need to be designed: systematic benchmarking the extraction accuracy and traceability of extractions to the sources' exact section, table or figure.

These limitations are compounded by problems specific to the electrochemical literature. Reporting conventions vary substantially across research groups and journals -- the same property can be expressed under different terminology, measured under different conditions, or reported at different levels of specificity. Electrolyte compositions are also more challenging to extract and parse than electrode materials. Cathode and Anode values are usually stated as single entities, whereas electrolyte formulations are inherently multi-component, comprising salts, solvents, and additives whose concentrations, ratios, and identities may be omitted, implied by reference to prior work, or described only partially. Linking such an entity to its measured properties is impossible without context from the rest of the document. Unit ambiguities are another source of error: conductivity values may appear in S/cm, mS/cm, or $\mu$S/cm without explicit conversion. These factors are even more pronounced for $\text{CO}_2$ electro-reduction, where reporting of cell configurations, reference electrodes, and product selectivity is far less standardized than in the battery community - so even a pipeline with strong NER~\cite{surveyonNamedEntityRecognition} and relation extraction can produce a structurally inconsistent database unless explicit mechanisms handle reporting heterogeneity at the document level, which makes cross-paper standardization and post-processing a prerequisite for any database intended for downstream machine learning tasks.

\begin{figure}[ht]
\centering
\includegraphics[width=0.8\linewidth]{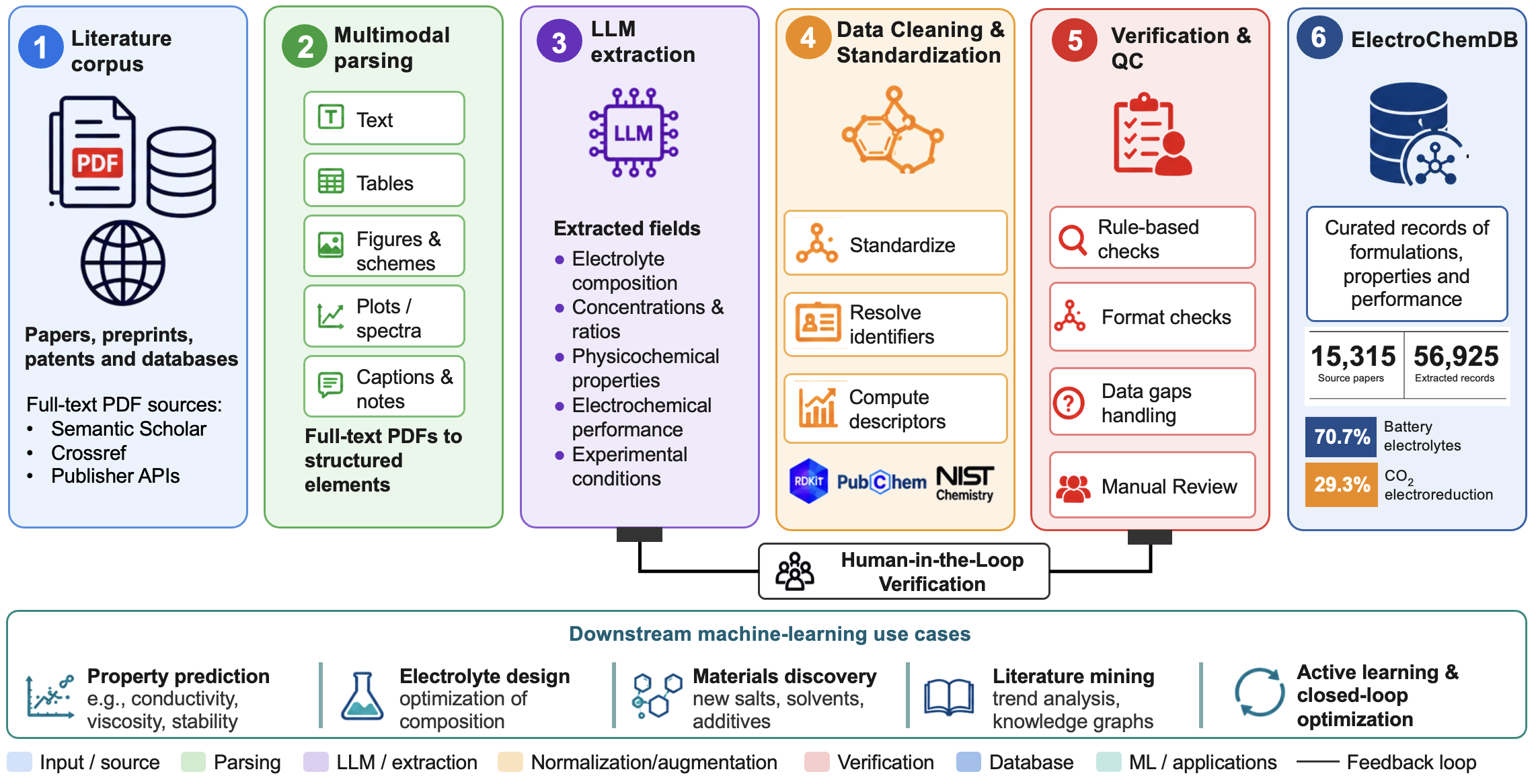}
\caption{
ElectroChemLLM workflow for constructing ElectroChemDB. Scientific PDFs are parsed into multimodal document representations comprising text, figures, and tables. Prompt-guided large language models aggregate information across modalities to extract complete electrolyte compositions (salts, solvents, additives, and concentrations), electrode and cell-chemistry information, physicochemical properties, and electrochemical performance metrics. Extracted records are standardised using external chemistry tools and databases, verified by domain experts, and consolidated into ElectroChemDB, enabling downstream machine-learning tasks including property prediction and materials discovery.
}
\label{fig:detailed_main_pipeline}
\end{figure}

To address these challenges, we introduce ElectroChemLLM, an end-to-end, multimodal, schema-driven pipeline that extracts complete electrolyte compositions, electrode and cell information, physicochemical properties, and electrochemical performance metrics from unstructured electrochemical literature. Our pipeline addresses each of the limitations in prior work: it reads text, tables, and figures jointly to overcome multimodal blindness, processes each paper as a whole to avoid context fragmentation, grounds every extracted value in its source location so that numerical hallucination becomes auditable, and standardises chemistry, units, and conditions so that heterogeneous reporting becomes a consistent, machine learning ready dataset. 

The pipeline combines a hybrid document parser---S2ORC-Grobid~\cite{lo-wang-2020-s2orc} for full text and section structure, Docling's~\cite{Docling} TableFormer for table reconstruction, and a fine-tuned YOLO~\cite{YOLO} detector for sub-panel-level figure localisation---with a vision-capable LLM-based summariser that translates each figure and table into a textual representation that travels alongside the body text. The full document context is then passed to an extraction LLM, which is constrained by an explicit JSON schema, designed in collaboration with domain experts, that defines target properties and their associated metadata. Attributing each generated statement to an identified source makes it verifiable~\cite{gao-etal-2023-enabling}, and selecting the supporting span before generating improves attribution accuracy and reduces human verification effort~\cite{slobodkin-etal-2024-attribute}. In materials extraction specifically, Polak et al.~\cite{Polak_2024} show that source attribution supresses fabricated entries. Applying the same principle to data curation: every extracted value is recorded with the section, table or figure it was taken from and check against it. This enables fabricated or misattributed values to be traced back to its source and flagged during verification. The resulting outputs are passed through an extensive chemistry-aware standardisation pipeline that reconciles heterogeneous chemical names, aliases, formulations, property labels, units, and measurement metadata into a consistent representation. This layer combines automated name resolution, canonical structure assignment, rule-based unit conversion, consistency checks, and targeted expert review for ambiguous or unresolved records.


This pipeline is deliberately modular and schema-driven, i.e., target properties can be added or restructured, new electrochemical subdomains incorporated, and the extraction LLM swapped out, all without modifying the rest of the architecture. We evaluate our pipeline's extractions on a manually extracted subset of 232 papers. ElectroChemLLM achieves an F-1 score of 0.97, outperforming both text-only baselines and prior extraction pipelines. Applied to a corpus of $\sim$15{,}000 articles spanning battery electrolytes and CO$_2$ electro-reduction, the pipeline produces ElectroChemDB---a standardized, queryable property database that we further use to train downstream ML models that predict cell-level specific capacity from extracted electrolyte, electrode, and test-condition fields, demonstrating a complete pathway from unstructured literature to predictive materials models. By converting literature that is largely locked in unstructured, multimodal documents into a standardised, provenance-linked and machine-readable resource, ElectroChemDB is intended to increase the value that can be derived from existing scientific data and to make it findable, interoperable and ready for AI-enabled discovery.  The rest of the paper is organised as follows: Section 2 outlines our design goals and the ElectrochemLLM pipeline, with component details in Appendix A. Section 3 includes our experimental setup, end-to-end extraction performance, database coverage analysis, and a downstream property-prediction case study (component-level and judge-based evaluation details are in Appendix B), followed by conclusions in Section \ref{sec13}.

\begin{table*}[t]
\centering
\footnotesize
\renewcommand{\arraystretch}{1.05}
\setlength{\tabcolsep}{2.4pt}

\begin{tabularx}{\linewidth}{
    @{}Xccccc>{\columncolor{green!15}}c@{}
}
\toprule
\textbf{Capability} &
\shortstack{\textbf{Direct}\\\textbf{LLM}} &
\textbf{CDE} &
\textbf{BDE} &
\textbf{MERMaid} &
\textbf{LLMB} &
\shortstack{\textbf{ElectroChem}\\\textbf{LLM}} \\
\midrule

Heterogeneous PDF ingestion
& \cmark & \pmark & \xmark & \cmark & \pmark & \cmark \\

Full-document contextual extraction
& \cmark & \pmark & \pmark & \cmark & \cmark & \cmark \\

Table extraction
& \pmark & \xmark & \xmark & \cmark & \xmark & \cmark \\

Figure understanding
& \pmark & \xmark & \xmark & \cmark & \cmark & \cmark \\


Structured property schema
& \cmark & \cmark & \cmark & \cmark & \cmark & \cmark \\

Chemical and unit normalisation
& \pmark & \pmark & \pmark & \pmark & \cmark & \cmark \\

Field-level source provenance
& \pmark & \pmark & \pmark & \pmark & \cmark & \cmark \\

Automated database construction
& \pmark & \cmark & \cmark & \cmark & \cmark & \cmark \\

Broad electrolyte-property scope
& \pmark & \pmark & \pmark & \xmark & \pmark & \cmark \\
\bottomrule
\end{tabularx}
\vspace{1pt}
{\scriptsize
\raggedright
\cmark{} indicates native support;
\pmark{} indicates partial, task-specific, or implementation-dependent support;
and \xmark{} indicates no support.
CDE and BDE denote ChemDataExtractor and BatteryDataExtractor, respectively.
\par}

\caption{Capability comparison of direct multimodal LLM extraction,
ChemDataExtractor 2.0~\cite{ChemDataExtractor,ChemDataExtractor2},
BatteryDataExtractor~\cite{huang2022batterydataextractor},
MERMaid~\cite{MERMaid}, LLMB~\cite{lee2026llmb}, and
\texttt{ElectroChemLLM}.}
\label{tab:electrochemllm_comparison}
\end{table*}

\section{Methods}
\label{methods}

To effectively extract information from full-text PDFs, with high fidelity and coverage across multiple modalities, we identified a set of desiderata that our framework must satisfy. First, since quantitative property values are distributed non-uniformly across the narrative text, tables, and figures, the parser must achieve 
“modality completeness”---i.e., accessing and collecting information from all the available content types---as any pipeline restricted to text or figures alone 
would omit a substantial fraction of reported data. The framework must therefore recover all available content types while preserving their internal structure, 
including the unit annotations that disambiguate values reported in varying conventions (e.g., conductivity in S/cm, mS/cm, or $\mu$S/cm; diffusion coefficients in 
$\text{cm}^2/\text{s}$ or $\text{m}^2/\text{s}$). Furthermore, the framework must demonstrate robustness when applied to $\text{CO}_2$ reduction literature.

Second, as a consequence of our first goal and given the nature of how information is distributed throughout the paper, it’s important to track the origin/grounding of the extracted text--in order to maintain transparency and enable validation/evaluation. Hence, our framework should annotate each extracted element to its source location (section, table, figure). 

Third, we want our framework to be able to unify information across multiple segments/modals--electrolyte entities are frequently introduced in one section, and characterized quantitatively across several pages/figures. Our framework must retain section hierarchy and cross-references, rather than emitting disconnected fragments---preserving hierarchical associations that sentence- and paragraph level frameworks often miss. 

Lastly, scalability and generalizability---our framework should be able to process tens of thousands of heterogeneous PDFs, spanning decades of typesetting conventions and publisher formats, within a compute budget compatible with downstream LLM inference costs. Beyond scale, it should also generalize across electrochemical subdomains---extending from battery electrolytes to adjacent use cases such as $\text{CO}_2$ electro-reduction---without requiring domain-specific re-engineering of the parsing or extraction stages.

Satisfying all four of these desiderata simultaneously is challenging--no single off-the-shelf parser recovers text, tables, and figures with the fidelity and grounding we need at scale. Hence, we designed a hybrid, multi-stage pipeline that decouples parsing along content-type boundaries, and delegates reasoning to a vision-capable LLM orchestrated through deterministic function calls.

Our parsing framework decouples extraction along content-type boundaries -- routing each modality through the tool best suited to handle it, and recombining the outputs into a unified document-level representation. Full article text is recovered via S2ORC-Grobid~\cite{lo-wang-2020-s2orc}, while tables and figures are processed in parallel by Docling's~\cite{Docling} TableFormer and a fine-tuned YOLO~\cite{YOLO} detector respectively. Each extracted element is tagged with its source location, and the parallel outputs are recombined into a single payload that is passed to the downstream vision-capable LLM stage. The pipeline relies only on open-source components and requires no publisher- or domain-specific tuning--supporting scalability and generalizability across electrochemical subdomains. We discuss each component separately in Appendix~\ref{app:electrochemllm_pipeline}.

\section{Results}
\label{results}

\subsection{Experimental Setup}
\label{subsec:experimental_Setup}

\paragraph{Benchmark.}
We evaluate \texttt{ElectroChemLLM} on a manually annotated benchmark of 232 papers spanning two electrochemical domains: 121 battery-electrolyte papers (526 electrolyte entries and 9,849 claims across 14 fields) and 111 CO$_2$ electroreduction papers (1,160 electrolyte entries). To ensure high annotation quality, we further investigate the interannotator agreement: a randomly selected subset of 50 records was independently annotated by two reviewers, yielding 95\% agreement and a Cohen’s \(\kappa\) of 0.90. For benchmarking the performance of our \texttt{ElectroChemLLM} pipeline against prior extraction pipelines and closed-source models, we use the subset of 121 battery-electrolyte papers.

\paragraph{Baselines \& Evaluation Metrics.}
We compare \texttt{ElectroChemLLM} against three categories of baselines: (a) established data-extraction frameworks, namely \texttt{ChemDataExtractor2}~\cite{ChemDataExtractor, ChemDataExtractor2}, configured with ten custom electrolyte models, and \texttt{BatteryDataExtractor}~\cite{huang2022batterydataextractor}, evaluated on its five supported properties; (b) the open-source model \texttt{ChemLLM-20B-SFT}~\cite{zhang2024chemllm}; and (c) the closed-source multimodal models \texttt{Gemini-2.5-Flash}~\cite{gemini25}, \texttt{Gemini-3.6-Flash}~\cite{gemini36}, and \texttt{GPT-4o}~\cite{openai2024gpt4ocard}, applied directly to the original PDFs. The direct-PDF models receive the same extraction instructions and output schema as \texttt{ElectroChemLLM}, but without document parsing or visual summarization. The existing frameworks and the open-source model are evaluated using text extracted with GROBID. The LLM baselines are evaluated on the complete 14-field schema, whereas the established frameworks are evaluated only on their supported fields. Therefore, their scores represent performance within their respective extraction scopes. \texttt{Gemini-3.5-Flash}~\cite{gemini35} compares the extracted records with the human-annotated ground truth through separate precision and recall passes. Precision, recall, and F1 are computed at the claim level, with matches required to belong to the same electrolyte formulation. Further implementation details and the complete judge prompt are provided in Appendix~\ref{app:results}.

\subsection{End-to-End Property Extraction Performance}
\label{subsec:main_results}
\begin{table*}[t]
\centering
\footnotesize
\setlength{\tabcolsep}{6pt}
\renewcommand{\arraystretch}{1.2}
\begin{tabular}{lccc}
\toprule
\textbf{Configuration} & \textbf{Precision} & \textbf{Recall} & \textbf{F1} \\
\midrule

\multicolumn{4}{l}{\textit{Prior Extraction Pipelines}} \\
\texttt{BatteryDataExtractor}$^{\dagger}$ & 0.18 & 0.14 & 0.16 \\
\texttt{ChemDataExtractor}$^{\dagger}$    & 0.37 & 0.28 & 0.32 \\

\midrule
\multicolumn{4}{l}{\textit{LLM Baselines}} \\
\texttt{ChemLLM-20B-SFT}                  & 0.57 & 0.45 & 0.50 \\
\texttt{GPT-4o} (Direct PDF)              & 0.75 & 0.44 & 0.55 \\
\texttt{Gemini-2.5-Flash} (Direct PDF)    & 0.43 & 0.85 & 0.57 \\
\texttt{Gemini-3.6-Flash} (Direct PDF)    & 0.66 & 0.74 & 0.70 \\
		
\midrule
\multicolumn{4}{l}{\textit{ElectroChemLLM}} \\
\texttt{GPT-4o}                           & 0.79 & 0.53 & 0.64 \\
\texttt{Gemini-2.5-Flash}                 & \textbf{0.94} & \textbf{0.98} & \textbf{0.97} \\

\bottomrule
\end{tabular}

\caption{End-to-end property-extraction performance on the 121-paper benchmark. Direct-PDF models receive the original PDFs without the document-processing and visual-summarization components of \texttt{ElectroChemLLM}. $^{\dagger}$Because the prior extraction pipelines support only specific property types, their results are calculated on the corresponding restricted property subsets; the LLM-based systems are evaluated on the complete output schema.}
\label{tab:main_results}
\end{table*}

Table~\ref{tab:main_results} presents the end-to-end property-extraction performance of \texttt{ElectroChemLLM} against the manually annotated benchmark. We report precision, recall, and F1 score for the complete pipeline and the selected baselines. As shown in Table~\ref{tab:main_results}, \texttt{ElectroChemLLM} with \texttt{Gemini-2.5-Flash} achieves the best overall performance, with an F1 score of 0.97, compared with 0.57 for direct-PDF extraction. The \texttt{GPT-4o} configuration achieves an F1 score of 0.64, while \texttt{ChemLLM-20B-SFT} reaches 0.50. The prior extraction pipelines obtain lower scores on their supported property subsets, highlighting the advantage of schema-guided LLM extraction. Property-level performance is consistently high (Table~\ref{tab:property-f1}), with a micro-average F1 of 0.97. The highest scores are for fields that are stated explicitly and in a consistent form (solvents, 99.55; salts, 98.61; cathode and anode, 98.47), whereas the weakest are overpotential (90.42), pH (90.91), additives (93.28) and total electrolyte composition (94.37), which are reported in heterogeneous or partial forms (additives, total composition) or can only be interpreted with context such as the reference electrode or current density (overpotential). Taken together, these results indicate that most of the gain over direct-PDF prompting comes from the pipeline rather than from the choice of LLM: with the same Gemini-2.5-Flash backbone, parsing, figure and table summarisation and schema-constrained, source-grounded extraction raise F1 from 0.57 to 0.97, whereas prior text-only pipelines reach only 0.16–0.32 on the properties they support. The smaller gain with GPT-4o (0.55 to 0.64) suggests that the backbone still matters, and the remaining errors concentrate on a mismatch between number of retrieved rows w.r.t. the ground truth. GPT-4o run variants are not able to match the ground truth's row granularity, consistently merging multiple reported conditions (such as a concentration or current-density series) into a single entry regardless of input pipeline, which caps precision and recall independent of parsing quality.

In throughput terms,Gemini-2.5-Flash with ElectroChemLLM pipeline processes roughly 2.3 papers per cent spent, compared with 0.53 for Gemini-3.6-flash direct-PDF, 0.41 for Gemini-2.5-Flash direct-PDF, and 0.12–0.23 for GPT-4o -- resulting in a 10- to 20-fold cost advantage alongside its higher F1 score.



We also evaluate the text-extraction, figure-localisation, and image-summarisation components individually. The figure-localisation model achieves 91\% accuracy, while the summarisation evaluation identifies coverage as the principal limitation. Coverage (1.8 out of 3) scores below accuracy (2.4) and faithfulness (2.7), i.e., the figure summaries are mostly correct but not exhaustive, so the dominant risk is that a value plotted in a figure is missed (lower recall) and not misreported (lower precision), consistent with the lower scores on fields such as overpotential that are often reported graphically. Full experimental details and results are reported in Appendix~\ref{app:results}.

\subsection{ElectroChemDB: Coverage and Distribution}
\begin{table*}[t]
\centering
\footnotesize
\setlength{\tabcolsep}{8pt}
\renewcommand{\arraystretch}{1.2}
\begin{tabular}{lc}
\toprule
\textbf{Statistic} & \textbf{Value} \\
\midrule
Total papers processed        & 232 \\
\quad Battery electrolytes    & 121 \\
\quad CO$_2$ electroreduction & 111 \\
\midrule
Total electrolyte entries     & 1,686 \\
\quad Battery electrolytes    & 526 \\
\quad CO$_2$ electroreduction & 1,160 \\
\midrule
Unique salts (names)          & 177 \\
Unique solvents (names)       & 110 \\
Unique additives (names)      & 152 \\
\midrule
\% entries with $\geq$1 quant. property & 98.0\% \\
\% entries with full composition        & 89.7\% \\
\bottomrule
\end{tabular}
\caption{\label{tab:db-stats}
Summary statistics for the manually validated benchmark subset of ElectroChemDB used for evaluation, comprising battery electrolyte and CO$_2$ electroreduction records with unit-standardised compositions and properties.}
\end{table*}

Having established the extraction performance of \texttt{ElectroChemLLM}, we characterise the benchmarked subset of \texttt{ElectroChemDB}. It contains 1,686 standardised electrolyte entries from 232 papers, including 121 battery-electrolyte and 111 CO$_2$ electroreduction papers (Table~\ref{tab:db-stats}). Overall, 89.7\% of entries contain complete composition information, and 98.0\% contain at least one quantitative property.

\begin{figure*}[ht]
    \centering
    \includegraphics[width=0.8\linewidth]{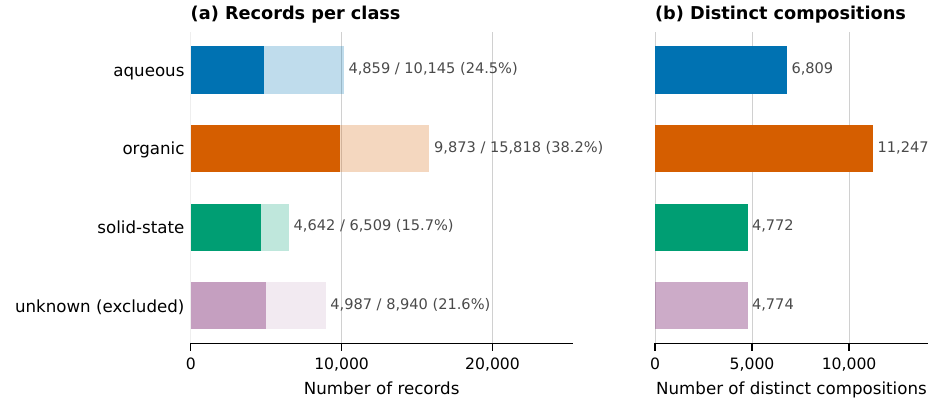}
    \caption{Distribution of records and distinct electrolyte compositions by
electrolyte class in the full battery-electrolyte corpus (41,144 records).
(a)Records per class: The solid segment counts records with at least one of
five properties (two electrolyte properties, ionic conductivity and viscosity, and three cell-level performance metrics that depend on both electrolyte and electrodes: cycle life, Coulombic efficiency and specific capacity) within its plausible physical range; the pale segment is
the remainder. Labels give usable/total records and the class's share of all
records. (b)Number of distinct electrolyte compositions per class. \emph{unknown} class shows unmapped compositions.}
\label{fig:be_classes}
\end{figure*}

The battery-electrolyte corpus comprises 41,412 records from around 11k articles: the
41,412 rows produced by the unit-conversion pipeline. We assign each record to one of four
electrolyte classes from its composition text (Fig.~\ref{fig:be_classes}).
Organic electrolytes dominate (38.4\%), followed by aqueous
(24.7\%) and solid-state (15.8\%) systems. The remaining 21\%
records could not be assigned a class and are excluded from all
subsequent analyses. Between 47.9\% (aqueous) and 71.3\% (solid-state) of each
class's records carry at least one property value that survives range
screening. The number of distinct compositions closely tracks the record count
for the three named classes (67.1--73.3\% of records are distinct), so the class
statistics below are not driven by a few formulations reported many times. The
unknown class is the exception, with only 55.1\% distinct compositions.  Organic and aqueous electrolytes together account for 63.1\% of all records, in line with the dominance of liquid electrolytes in the battery literature, whereas solid-state systems are a minority of records (15.8\%) but have the highest share with a usable property value (71.3\%), plausibly because ionic conductivity is the headline figure of merit in that literature.


\paragraph{Property Coverage and Chemical Diversity.}

\begin{figure*}[ht]
    \centering
    \includegraphics[width=0.80\linewidth]{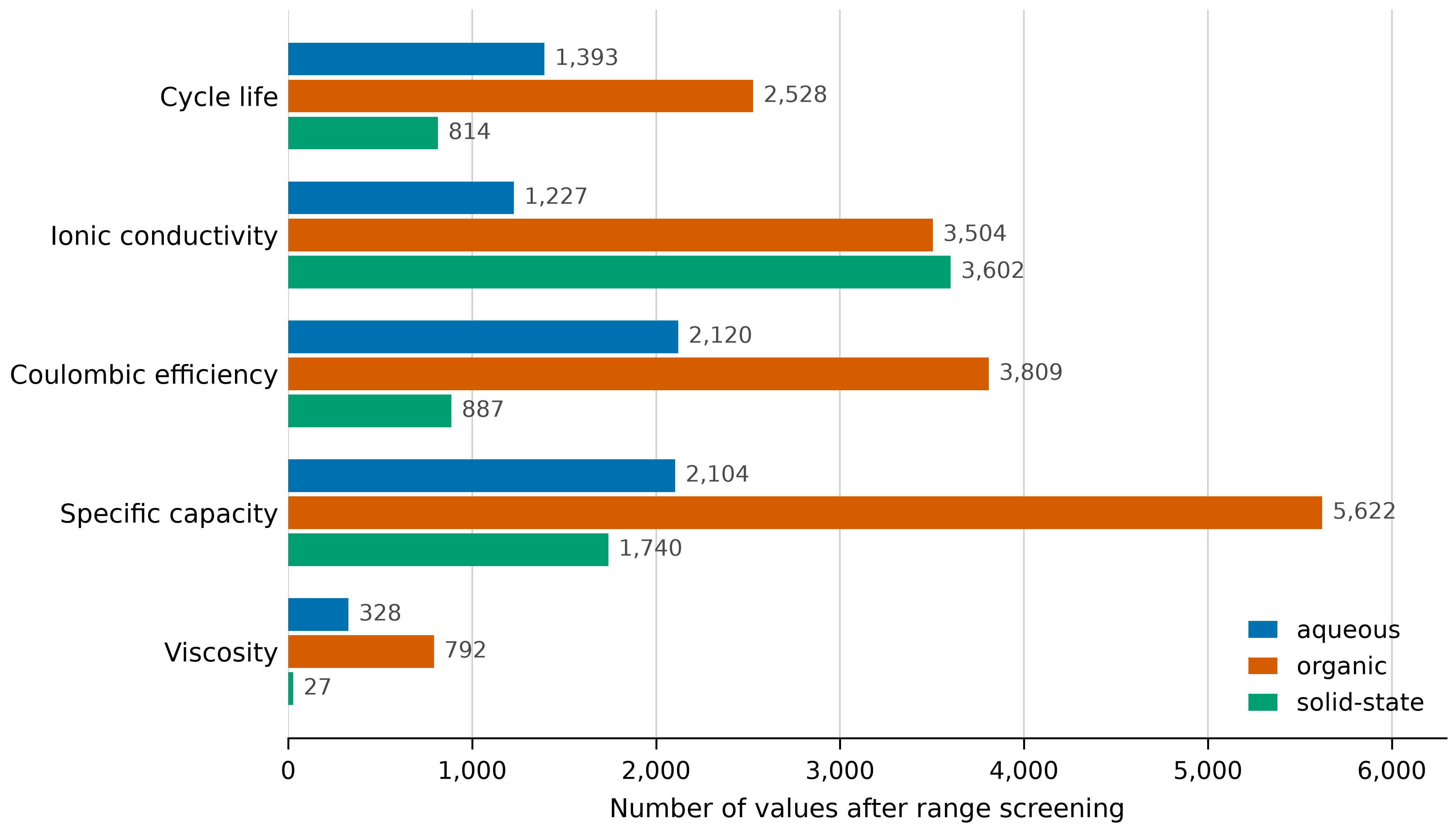}
    \caption{Number of values per property and electrolyte class in the full
battery-electrolyte corpus, after removing values outside each property's
plausible physical range. Excludes unknown class.}
    \label{fig:data_coverage_per_axis}
\end{figure*}

Property coverage is uneven across classes (Fig.~\ref{fig:data_coverage_per_axis}).
Organic electrolytes have the most values for every property except ionic
conductivity, where solid-state systems are slightly ahead (3,602 vs.\ 3,504 values). Specific capacity is the best-covered property, with up to 5,622
values for organic electrolytes. Viscosity is the sparsest, with only 27
solid-state values, so viscosity comparisons involving solid-state systems
should be read with caution.

\begin{figure}[t]
\centering
\includegraphics[width=0.8\linewidth]{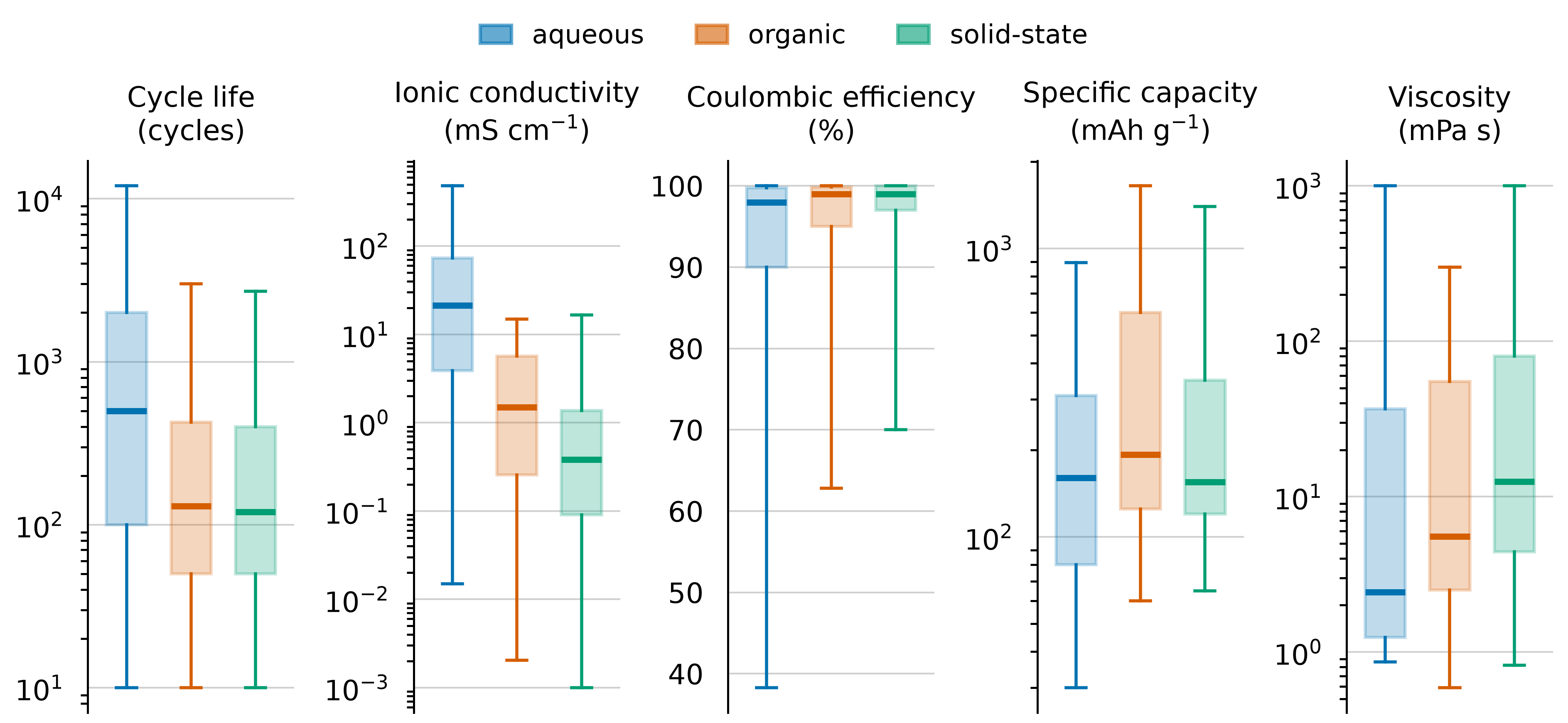}
\caption{Distributions of cycle life, ionic conductivity, Coulombic efficiency,
specific capacity and viscosity by electrolyte class in the full
battery-electrolyte corpus. Boxes span the interquartile range, the thick line
marks the median, and whiskers extend to the 5th and 95th percentiles. All axes
are logarithmic except Coulombic efficiency.}
\label{fig:property_distributions_boxplots}
\end{figure}

Figure~\ref{fig:property_distributions_boxplots} shows the distribution of each property by
class in physical units. First, ionic conductivity separates the classes most clearly. Aqueous electrolytes have the highest median (21.3~mS\,cm$^{-1}$;
interquartile range 3.9--73.5), followed by organic (1.5~mS\,cm$^{-1}$) and
solid-state systems (0.38~mS\,cm$^{-1}$). This ordering matches the literature:
moderately concentrated acidic and alkaline aqueous electrolytes (e.g., KOH, H$_2$SO$_4$) routinely exceed 100~mS\,cm$^{-1}$, whereas
ultra-concentrated water-in-salt formulations fall to about 10~mS\,cm$^{-1}$~\cite{suo2015water}, and most polymer and oxide solid electrolytes fall well below 1~mS\,cm$^{-1}$ at room temperature,
with only a few sulfide conductors reaching liquid-like
values (~\cite{kamaya2011lithium}, ~\cite{Famprikis2019}). The organic median lies below the
${\sim}10$~mS\,cm$^{-1}$ typical of carbonate electrolytes \cite{xu2004nonaqueous,xu2014electrolytes}
because the class also contains ionic-liquid, gel, and highly concentrated
formulations and measurements below room temperature. Second, aqueous systems
report the longest and most widely spread cycle life (median 500 cycles;
interquartile range 100--2,000), compared with 130 and 120 cycles for organic
and solid-state electrolytes. This likely reflects the many aqueous zinc and
high-rate cells cycled for thousands of cycles. Because cycle life depends on rate, depth of discharge, and the end-of-life criterion, which differ between papers, these values are indicative rather than strictly comparable. Third, Coulombic efficiency
medians are similar across classes (98--99\%), but the aqueous distribution has
a much longer low-efficiency tail (5th percentile 38\%, against 63\% for
organic and 70\% for solid-state electrolytes). This is consistent with the
narrow electrochemical stability window of water, where hydrogen evolution and electrode corrosion consume charge. The wide ranges on logarithmic axes reflect the diversity of
chemistries and measurement conditions, and the number of values behind each
box is given in Fig.~\ref{fig:data_coverage_per_axis}.

While the extracted data show trends consistent with the electrochemical literature, comparisons across electrolyte classes need some caution. Properties such as ionic conductivity and cycle life depend on composition, temperature, and testing conditions, which vary considerably between studies. The distributions also reflect what has been reported in the literature, including a bias toward successful results, rather than the full range of performance possible for each electrolyte class.

\paragraph{Source Provenance.}

\begin{figure*}[!ht]
    \centering
    \includegraphics[width=0.8\linewidth]{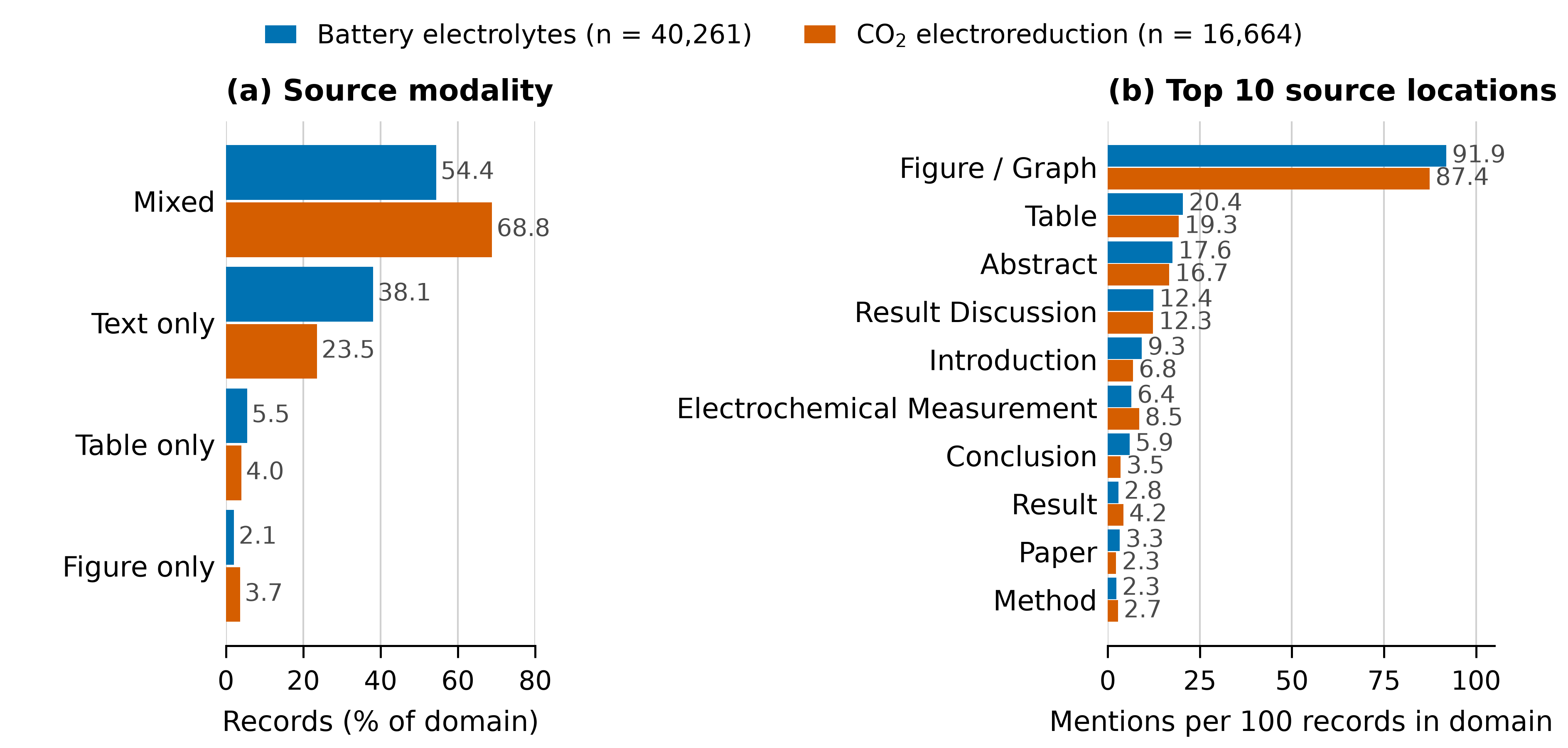}

    \caption{Source provenance of ElectroChemDB records from the multimodal extraction run: 56,925 records from 15k papers, split into battery electrolytes ($n = 40{,}261$) and CO$_2$ electroreduction ($n = 16{,}664$). (a) Distribution of records by source modality, as a percentage of each domain's records. Mixed records draw their fields from more than one of text, tables and figures; table only and figure only records come entirely from a single table or figure. (b) The ten most frequently cited sources. Labels were normalized; numbered tables and figures were collapsed into Table and Figure / Graph.}
    \label{fig:source_provenance}
\end{figure*}

Each database entry retains its source modality and document location. Of all the successfully source tagged entries across all 15K papers, 66.2\% draw on at least one table or figure: 58.6\% combine text with tables or figures, and a further 7.6\% come exclusively from a table
(5.0\%) or a figure (2.6\%). As shown in Fig.~\ref{fig:source_provenance}, figures and graphs are by far the most frequently cited sources, followed by tables and abstracts. This highlights the importance of capturing information beyond the main text, particularly for properties that are primarity reported through plots and tables.

The benefit of this approach is also reflected in the benchmark results, where ElectroChemLLM achieves an F1 score of 0.97, outperforming both text-only extraction pipelines and direct-PDF LLM extraction. Together, these results demonstrate the value of combining text, tables, and figures for large-scale electrochemical data extraction.

\subsection{Downstream Application: Property Prediction}
\label{subsec:ml_pipeline}
\begin{figure}[H]
\centering
\includegraphics[width=0.9\linewidth,height=0.75\textheight,keepaspectratio]{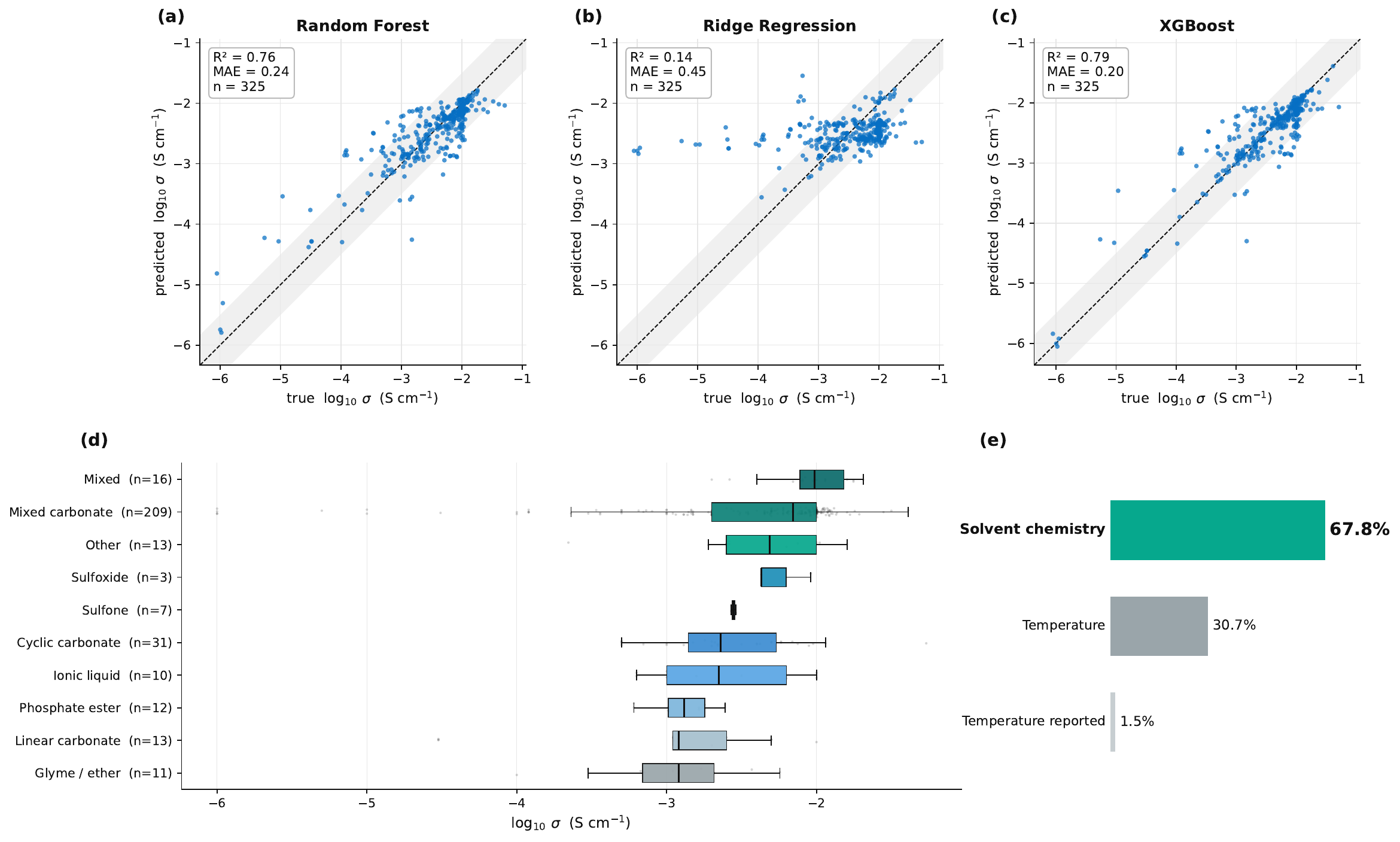}
\caption{ML readiness of the Li-ion electrolyte slice of ElectroChemDB, evaluated on ionic conductivity. (a--c) In-sample parity plots for (a) Random Forest, (b) Ridge regression, and (c) XGBoost on concentration-weighted ECFP composition features; the dashed line indicates $y=x$, and the shaded band corresponds to $\pm 0.5$ $\log_{10}$ units. (d) $\log_{10}\sigma$ by solvent class ($n \geq 3$, ordered by descending median). (e) Random Forest feature importance aggregated by descriptor type across the top 20 features.}
\label{fig:mlcomponent}
\end{figure}

\subsubsection{ML utility of ElectroChemDB.} A core motivation for ElectroChemDB is that the curated, standardised entries should be directly consumable by downstream ML pipelines for materials discovery. Unlocking the value of the published literature for AI-enabled discovery requires data that are standardised, linked to their measurement context, and ready for use without further manual curation. We test this with a case study that predicts cell-level specific capacity, the best-covered of the five screened properties (Fig.~\ref{fig:data_coverage_per_axis}), for Li-ion and Li-metal cells separately. The case study asks two questions: whether models trained directly on records extracted by ElectroChemLLM learn a non-trivial relationship between a cell’s description and its measured capacity, and which parts of that description carry the signal. All results below are out-of-fold predictions for entries the model did not see during training.

\subsubsection{Dataset and representation.}

We use the Li-ion and Li-metal battery records of ElectroChemDB, split by the negative electrode (graphite or Si versus Li foil) and modelled separately. Extracted formulations are first converted to a common molar-concentration basis (mol~L$^{-1}$), while the original reported composition and component-specific conversion status are retained; the conversion rules, assumptions and coverage statistics are given in Appendix~\ref{app:composition_normalization}.

After strict, rule-based cleaning, which removes every row
that has a component that cannot be mapped to a validated
SMILES, the sets contain 970 Li-ion entries (480 papers,
191 distinct formulations) and 1,116 Li-metal entries
(580 papers, 228 distinct formulations). The prediction
target is the base-10 logarithm of the reported specific
capacity, which spans roughly three orders of magnitude
(Fig.~\ref{fig:mlcomponent}). These records are LLM-extracted and
have no human-annotated reference, so the case study tests
whether the database can be used as delivered, not whether
it reproduces curated labels.

Each entry is described by three groups of inputs.
(i) \textit{Electrolyte composition}: a 2048-bit Morgan/ECFP
fingerprint (radius 2; Rogers and Hahn 2010) is computed
for every component, and the per-component fingerprints
are aggregated within role by concentration-weighted
averaging, with additives included as extra components.
The resulting vector has the same length regardless of
recipe size, so multi-component electrolytes (e.g.,
EC + DMC + DEC + EMC + LiPF$_6$) and single-component
formulations share one feature space. (ii) \textit{Electrode information}: total salt
concentration, number of additives, and anode and cathode
class (31 features). (iii) \textit{Test conditions}: current density, C-rate,
areal capacity, cycle number, cell configuration and
voltage window (18 features). Groups (ii) and (iii) are stored alongside each formulation
in ElectroChemDB, which is what allows them to be used here
without further extraction.

\subsubsection{Models and validation.} 

We compare three standard regression models, Random Forest (RF),
Ridge Regression and XGBoost
\cite{breiman2001random,hoerl1970ridge,chen2016xgboost},
with a pretrained graph neural network
(GNN)~\cite{yang2025self,rampavsek2022recipe}
that receives the same electrode and test-condition features
at its prediction head. The classical models use identical inputs. Performance is assessed only out of fold, with 5-fold
cross-validation over shuffled rows, so that every point in
Fig.~\ref{fig:mlcomponent} is a prediction for an entry the model did
not see during training; in-sample fits, which overstate
performance, are not used. The classical models are repeated
with five different shuffles; the GNN is run once, on the
folds of the first repeat, which is the repeat plotted.
We report the coefficient of determination $R^2$ and the
mean absolute error (MAE) in $\log_{10}$ units (definitions
in Section~5.1). Because the target is logarithmic, an MAE
of 0.17 corresponds to a typical error of about a factor
of 1.5 in the capacity itself. Two further tests probe what the models have learned: an
electrolyte-only variant, which isolates how much signal
comes from the formulation, and a grouped split in which
every formulation is confined to a single fold, which
tests generalisation to unseen electrolytes.

\subsubsection{Predictive performance.} 

\noindent\textbf{Cross-validated accuracy.}
With electrolyte, electrode and test-condition inputs,
XGBoost and RF reach $R^2 = 0.735$ and $0.726$ on Li-ion
(MAE = 0.173 in $\log_{10}$ units), ahead of the GNN
($R^2 = 0.645$, MAE = 0.201) and Ridge
($R^2 = 0.562$, MAE = 0.247)
(Fig.~\ref{fig:mlcomponent}). On Li-metal, XGBoost, RF and the
GNN reach $R^2 = 0.500$, $0.497$ and $0.469$
(MAE = 0.152--0.157), and Ridge $0.352$
(MAE = 0.197). Averaged over the five shuffles,
RF and XGBoost give $R^2 = 0.71$ and $0.72$
on Li-ion and $0.51$ and $0.52$ on Li-metal,
so the repeat plotted in Fig.~\ref{fig:mlcomponent}
is representative. On Li-ion cells (top row of Fig.~\ref{fig:mlcomponent}),
the tree-based models follow the parity line over
most of the range, while Ridge shows visibly wider
scatter. On Li-metal cells (bottom row), the scatter
is larger for every model and predictions are pulled
toward the centre of the distribution: the highest
reported capacities ($\log_{10}$ above about 3.3)
are underpredicted and the lowest are overpredicted.
Li-metal is also harder to rank ($R^2$ about 0.50
versus 0.74 for Li-ion) even though its absolute
error is slightly smaller (MAE about 0.15 versus
0.17), which implies a narrower spread of capacities,
so that the same error explains less of the variance.
The tree-based models outperform Ridge by about
0.15--0.17 in $R^2$ on both chemistries, consistent
with a non-linear dependence of capacity on the
interplay between electrode, current density and
cycling conditions. The pretrained GNN does not
outperform the tree-based models (0.645 versus
0.735 on Li-ion; 0.469 versus 0.500 on Li-metal);
with only 191 and 228 distinct formulations,
a plausible reason is that there is little for
a learned electrolyte representation to exploit.

\begin{figure}[H]
\centering
\includegraphics[width=\linewidth]{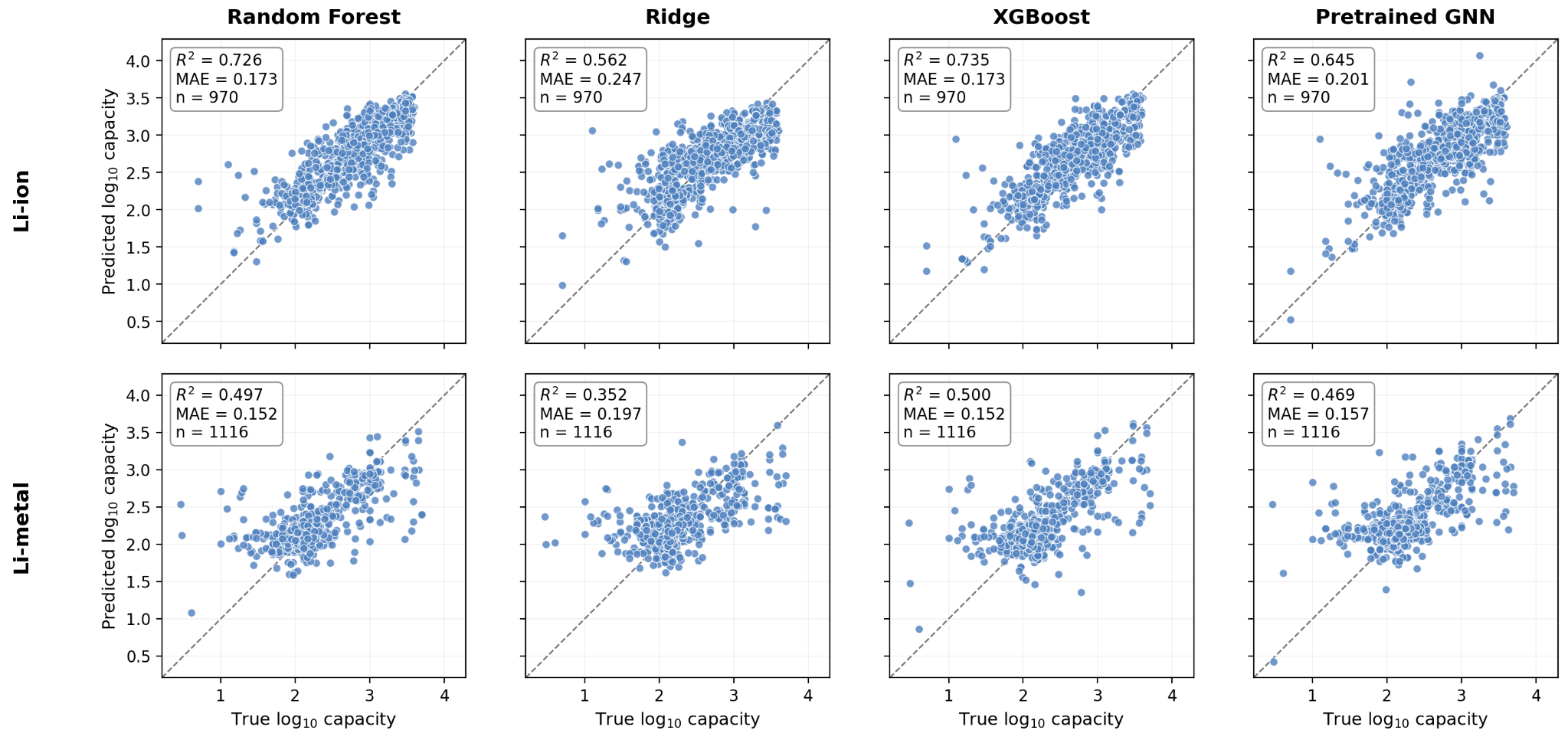}
\caption{Out-of-fold predictions of $\log_{10}$ specific capacity for Li-ion (top) and Li-metal (bottom) cells. Columns show Random Forest, Ridge, XGBoost, and a pretrained GNN using electrolyte, electrode, and test-condition inputs. Each panel reports $R^2$ and MAE (in $\log_{10}$ units) for the first of five cross-validation repeats, with $n$ the number of entries; dashed lines indicate perfect agreement. Both axes show $\log_{10}$ of the reported specific capacity; each point is one extracted entry, predicted by a model that did not see it during training. GNN, pretrained graph neural network.}
\label{fig:augmentation}
\end{figure}

\subsubsection{Effects of input features and generalisation.}

Using electrolyte composition alone, the models achieve $R^2 = 0.17$--$0.38$ on Li-ion and $0.20$--$0.28$ on Li-metal cells. Further adding the electrode and test-condition fields raises the best scores to $0.735$ and $0.500$ respectively. Since a reported capacity depends on the active materials, current density, voltage window and cycle number at least as much as
on the electrolyte, retaining this information adds value to ElectroChemDB for downstream property prediction. 

The random cross-validation results aloow the same electrolyte formulation to apprear in both training and test sets. When formulations are kept separate across folds, we observe limited predictive performance, with electrolyte-only $R^2$ ranging from $-0.07$ and $0.04$. This reflects the diversity of electrolyte compositions and reported property ranges across the literature, suggesting that substatially more data covering a wider range of formulations will be needed to generalise to unseen electrolytes.

Overall, the results show that ElectroChemDB provides standardised, machine-learning-ready data without requiring additional manual curation beyond rule-based filtering. The case study also highlights the value of extracting electrolyte compositions together with electrode information
and experimental conditions.

\section{Conclusion}\label{sec13}
We introduced ElectroChemLLM — a multimodal, end-to-end pipeline for extracting electrolyte chemical entities and their physicochemical properties from unstructured scientific literature. By combining hybrid document parsing with vision-capable LLM orchestration and a schema-driven extraction layer, the pipeline jointly addresses the three main limitations of prior work in this space - multimodal blindness, context fragmentation, and numerical hallucination. On a manually annotated benchmark of 232 papers, ElectroChemLLM achieves an F-1 score of 0.97 - substantially outperforming both text-only baselines and prior extraction pipelines. We applied the pipeline to a corpus of 15,000 articles spanning battery electrolytes and $CO_2$ electroreduction, to construct ElectroChemDB — a standardised, queryable property database that we further use to train downstream ML models that predict cell-level specific capacity from extracted electrolyte, electrode
and test-condition fields (out-of-fold $R^2$ up to $0.735$ for Li-ion and $0.500$ for Li-metal cells) illustrating a complete pathway from unstructured literature to predictive materials models.

ElectroChemLLM pipeline is modular by design, allowing new properties, electrochemical domains, and LLM backends to be incorporated without redesigning the overall workflow, making the pipeline adaptable to future data extraction tasks with minimal changes.


\backmatter

\bmhead{Data \& Code Availability}
The source code and benchmark dataset compiled during this study, including our prompts, pipeline modules, $\text{CO}_2$ reduction and Battery electrolytes properties, will be released publicly on GitHub. 






\bmhead{Acknowledgment}

This work was supported by the National Science Foundation under Grant Nos. NSF CSSI-2514002 and NSF CBET 2505289, and by seed funding from Stony Brook University Institute of Advanced Computational Science. We thank George Mirgorodskiy and Maggie Li for their assistance with battery-papers benchmarking and Ray Zhang (St. George’s School, Vancouver, Canada), Yuanxi “Cici” Yao (Ho Chi Minh City, Vietnam) and Albert Huang (Saigon South International School) for their assistance with 
CO$_2$ER benchmarking.











\bibliography{sn-bibliography}

\begin{appendices}
\section{ElectroChemLLM Pipeline}
\label{app:electrochemllm_pipeline}

\subsection{Component A: Knowledge Base Creation}
\label{subsec:knowledge_base_creation}

Our parsing pipeline begins with targeted literature acquisition from high-impact journals and repositories. We start by querying Semantic Scholar~\cite{semanticscholar} and CrossRef~\cite{Crossref} using a curated set of electrochemistry-specific keywords, filtering for articles that are either open-access or have a direct PDF link available. The resulting candidate list is then used to fetch PDFs from the respective publishers--each of which exposes its content slightly differently, requiring publisher-specific handling at the download stage. At the end of this step, we are left with a total of ~15K papers--~10K covering battery electrolytes and ~5K covering $\text{CO}_2$ electro-reduction.
\subsection{Component B: Document Parser}
\label{subsec:document_parser}
Once the relevant PDFs have been acquired, the next step is to extract their content in a form that preserves both textual and visual information. As outlined in Section~\ref{methods}, no single off-the-shelf parser recovers all modalities with the fidelity we need--hence, so we decouple extraction along content-type boundaries and route each modality through a dedicated tool. Full text and captions are handled by S2ORC-Grobid(Section~\ref{subsubsec:full_text}), tables are reconstructed using Docling's TableFormer model(Section~\ref{subsubsec:table_extraction}), and figures are localised using a fine-tuned YOLO detector(Section~\ref{subsubsec:figure_extraction}). The outputs of these three streams are then merged into a unified document-level representation, which serves as the input to the downstream LLM extraction stage.

\subsubsection{Full Text Extraction with S2ORC-Grobid}
\label{subsubsec:full_text}

For the textual backbone of each paper, we use the S2ORC doc2json parser--a utility from Allen AI that converts scientific PDFs into a semi-structured, machine readable JSON format. Under the hood, S2ORC relies on the Grobid ML library to first convert raw PDFs into a structured TEI-XML format, which is then passed through a custom grobid2jsn step to extract paper components into a final json output. We run this stage in a locally configured Python environment (setup details available in our GitHub repository).

\begin{table*}[t]
\centering
\footnotesize
\setlength{\tabcolsep}{4pt}
\renewcommand{\arraystretch}{1.2}
\begin{tabularx}{\columnwidth}{l X}
\toprule
\textbf{Field} & \textbf{Content} \\
\midrule
Paper ID & \texttt{c5ee03874j.grobid.tei} \\
\midrule
\multicolumn{2}{l}{\textit{\textbf{metadata}}} \\
Title & Cesium-containing triple cation perovskite solar cells: improved stability [...] \\
Authors & Michael Saliba, Taisuke Matsui, Ji-Youn Seo, [...] \\
DOI & 10.1039/c5ee03874j \\
Publisher & Royal Society of Chemistry (RSC) \\
Published date & 2016-03-16 \\
Abstract & Today's best perovskite solar cells use a mixture of formamidinium and methylammonium [...] \\
\midrule
\multicolumn{2}{l}{\textit{\textbf{Text (section-keyed})}} \\
Introduction & Perovskite solar cells have attracted enormous interest in recent years [...] \\
Results and discussion & We investigated triple cation perovskites of the generic form [...] \\
Film characterisation & In Fig.~1a, we show X-ray diffraction (XRD) data for C$_x$M with [...] \\
Device characterisation & As illustrated in the cross-sectional SEM images in Fig.~3, the solar cell [...] \\
Conclusion & This work, for the first time, used a mixture of a triple Cs/MA/FA cation [...] \\
\bottomrule
\end{tabularx}
\caption{\label{tab:s2orc-output}
Example S2ORC-Grobid output structure for a representative paper. The parser recovers structured bibliographic metadata, abstract, and section-keyed body text with inline figure/table references preserved. Visual content and table structure are not recovered at this stage.
}
\end{table*}

The output, as shown in Table~\ref{tab:s2orc-output}, gives us a structured, section-by-section representation of the paper--bibliographic metadata, the abstract, and the full body text organized by section heading--with inline references to figures and tables preserved in-place. This addresses our document-level context desideratum. However, S2ORC does not recover the visual content of figures or the internal structure of tables--for those, we turn to Docling's TableFormer and a fine-tuned YOLO detector described in Section~\ref{subsubsec:table_extraction} and~\ref{subsubsec:figure_extraction}.

\subsubsection{Table Structure Recognition with Docling and TableFormer}
\label{subsubsec:table_extraction}

Since S2ORC does not recover the internal structure of tables, we use the Docling framework to handle this. Docling~\cite{Docling} is an open-source document processing package powered by a suite of specialized models for layout analysis and table structure recognition. We use it specifically for tables--the layout analysis model (an object detector trained on DocLayNet~\cite{Pfitzmann_2022}) first localises table regions on each page, and the TableFormer model then takes these cropped table images and predicts their logical row and column structure, including merged cells and nested headers.
The output of this stage is a structured representation of each table that preserves the spatial relationships between cells--which, as noted in our desiderata, is essential for keeping property-entity associations intact. Alongside the structured form, we store the tables in structured CSV formats and augment the table data with the Grobid JSON output.

\subsubsection{Figure Localisation with a Fine-Tuned YOLO Detector}
\label{subsubsec:figure_extraction}

\begin{figure}
\centering
\includegraphics[width=1\linewidth]{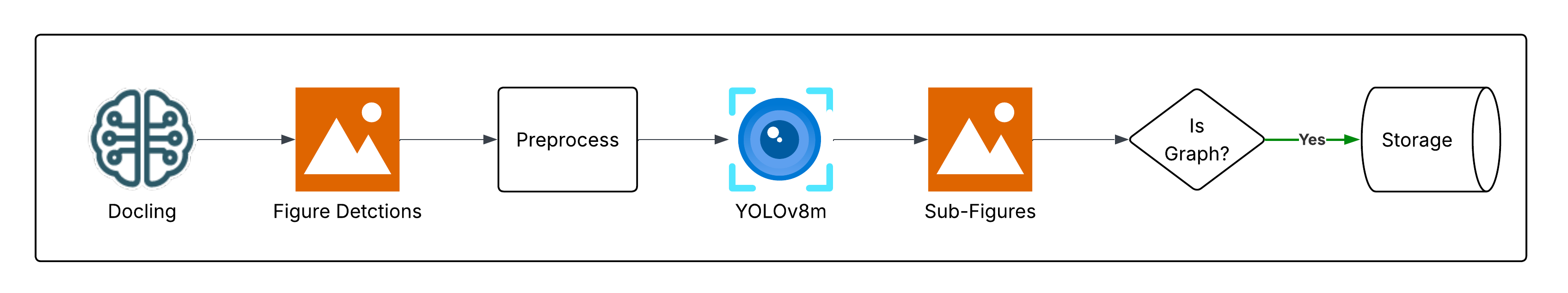}
\caption{
Sub Figure extraction via YOLOv8 Processing.
}
\label{fig:yolo_processing}
\end{figure}

\begin{figure}
\centering
\includegraphics[width=1\linewidth]{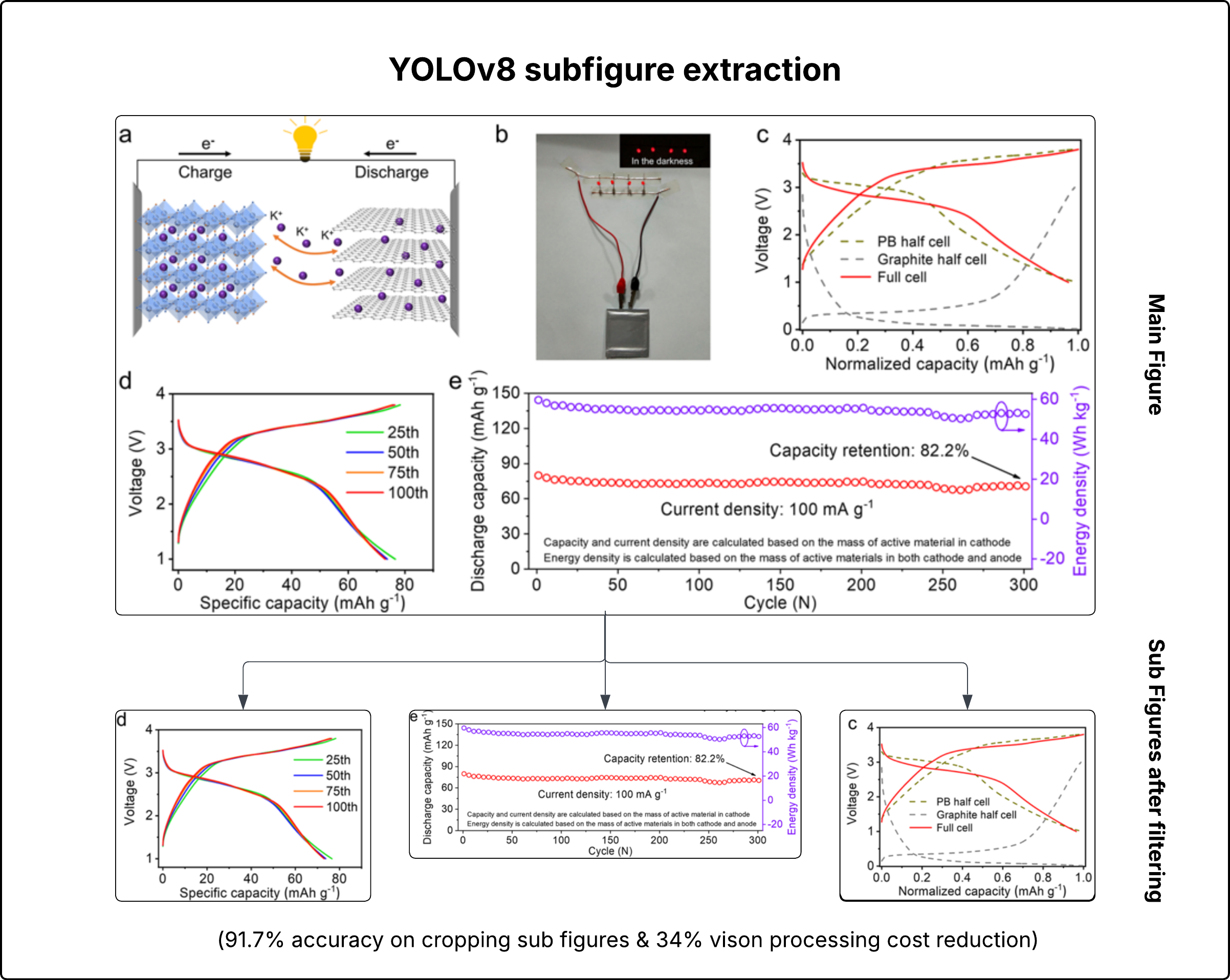}
\caption{
YOLOv8 SubFigure Extraction to isolate individual sub-panels from a composite image.
}
\label{fig:subfigure_extraction}
\end{figure}

For figure extraction, we build on Docling's layout analysis model--an object detector originally trained on DocLayNet that predicts bounding boxes and classes for page elements such as figures, tables, captions, and text blocks. While the base model performs reasonably well on general scientific documents, it struggles with a pattern that is especially common in electrochemistry literature--composite figures made up of multiple sub-panels clustered together under a single caption, where each sub-panel reports a different property or experimental condition. Figure~\ref{fig:subfigure_extraction} for example, shows a composite figure comprising charge/discharge voltage profiles, capacity versus cycle number, and energy density curves reported side by side under a single caption. When the base detector treats such a cluster as a single figure, the downstream LLM receives one merged image in which individual property plots are difficult to isolate, leading to mis-associations between property values and their axes, units, or sample labels. Accurate sub-panel localisation is therefore critical for preserving the one-to-one mapping between a plotted quantity and its extracted value.

To address this, we introduce a separate YOLO detector, fine-tuned on a manually curated set of electrochemistry figures. The training set comprises 496 images extracted through docling, augmented using standard computer vision techniques (grayscale conversion, hue, saturation, exposure, and noise perturbation) to yield 1,296 training images in total, with 2 label classes: graphs and diagrams. The YOLO detector is applied downstream of Docling's layout model--once a composite figure region has been identified on the page, the YOLO detector operates within that region to isolate individual sub-panels(as shown in fig~\ref{fig:subfigure_extraction}). Each sub-panel is then cropped, Base64-encoded, and passed downstream alongside the structured table outputs from~\ref{subsubsec:table_extraction}. As with tables, retaining sub-panels as individual images (rather than attempting to parse their content at this stage) lets the downstream vision-capable LLM reason directly over each plot and its associated axes, legends, and in-plot annotations.

At the end of this stage, we have three complementary streams of extracted content for each paper: (i) section-level text and metadata from S2ORC-Grobid, (ii) structured, cell-level table representations from Docling's TableFormer, and (iii) sub-panel-level figure crops from our fine-tuned YOLO detector. Before these streams can be merged into a unified document-level payload, the figure crops need to be converted into a form that can be reasoned over alongside the text--which brings us to our next component, the Image Summarizer.

\subsection{Component C: Image and Table Summarizer}
\label{subsec:image_summarizer}
The outputs from our \texttt{Document Parser} provides, per paper, a set of individual figure crops and structured table representations alongside the full text. Before these can be merged into a unified payload for downstream extraction, each visual artifact is passed through an image summarizer: a vision-capable LLM (Gemini 2.5 Flash, processing figures in batches of ten per API call) that generates a detailed textual description of the figures in isolation.
The motivation for this step is granularity of attention--when a large multimodal payload containing many figures and tables is passed to a single LLM call, individual visual details (axis labels, unit annotations, sub-panel legends, specific cell values) are often skimmed over or missed entirely, since the model distributes its attention across the full context. By processing each image and table individually, we ensure every visual artifact receives the model's full attention, and all fine-grained details--sample names, reported values, measurement conditions, units--are captured in the resulting summary. 

\begin{table*}[t]
\centering
\small
\renewcommand{\arraystretch}{1.3}
\begin{tabularx}{\textwidth}{c X}
\toprule
\textbf{Input Figure Crop} & \textbf{Summariser Output (JSON)} \\
\midrule
\raisebox{-\totalheight}{\includegraphics[width=0.35\textwidth]{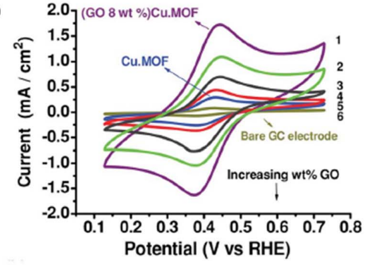}}
&
\ttfamily\scriptsize
\{\par
\quad "summary": "A cyclic voltammetry (CV) plot showing the electrochemical performance of different electrodes [...]",\par
\quad "key\_findings": [\par
\quad\quad "The bare GC electrode exhibits negligible current response.",\par
\quad\quad "The Cu.MOF electrode shows a pair of redox peaks [...]",\par
\quad\quad "The (GO 8 wt\%)Cu.MOF electrode demonstrates significantly enhanced current density [...]",\par
\quad\quad "Increasing the weight percentage of GO generally leads to an increase in current response [...]",\par
\quad\quad "[...]"\par
\quad ]\par
\}
\\
\bottomrule
\end{tabularx}
\caption{\label{tab:summariser-example}
Example input–output pair from the image summariser. The left column shows a figure crop extracted by the fine-tuned YOLO detector~\ref{subsubsec:figure_extraction}; the right column shows the corresponding JSON summary produced by the vision-capable LLM, capturing figure-level context, key quantitative findings, and reference metadata.
}
\end{table*}

An example input-output pair from this step is shown in Table~\ref{tab:summariser-example}, where a cyclic voltammetry figure is translated into a structured JSON summary capturing both high-level context and specific quantitative findings. These summaries are text, which means they can be merged directly with the body text from S2ORC-Grobid into a single document-level context for the downstream extraction stage, without running into the context-length or multimodal-payload-size limits of a typical LLM call.

\subsection{Component D: LLM-Based Property Extraction}
\label{subsec:llm_based_extraction}

The next step of the pipeline is to convert the assembled document-level context into structured, queryable property data--the actual target of this work. With body text from the \texttt{document parser} and structured summaries of figures and tables from the \texttt{image summarizer}  now merged into a single JSON payload, we can hand the full paper to an LLM in text form and ask it to identify electrolyte entities and their associated physicochemical properties. We break this step into two parts: Section~\ref{subsubsec:schemas} defining schemas (i.e. what to extract), and Section~\ref{subsec:callAPI} executing the extraction via a structured API call.

\subsubsection{Property Selection}
\label{subsubsec:schemas}

Electrochemical literature reports a wide range of properties, but not all are equally relevant for downstream materials discovery--and reporting conventions vary considerably across papers, so a naive "extract everything" approach tends to produce noisy, inconsistent outputs. We therefore worked with material science experts to define a focused set of target properties that are both scientifically meaningful for electrolyte design and commonly reported in the literature.

\begin{table*}[t]
\centering
\footnotesize
\setlength{\tabcolsep}{4pt}
\renewcommand{\arraystretch}{1.2}
\begin{tabularx}{\columnwidth}{l X}
\toprule
\textbf{Property} & \textbf{Captured Fields} \\
\midrule
\multicolumn{2}{l}{\textit{\textbf{Electrolyte Composition}}} \\
Total composition & Full electrolyte description \\
Salt(s) & Name, concentration, unit \\
Solvent(s) & Name, ratio \\
Additive(s) & Name, concentration, unit \\
\midrule
\multicolumn{2}{l}{\textit{\textbf{Cell Information}}} \\
Cathode & Material \\
Anode & Material \\
\midrule
\multicolumn{2}{l}{\textit{\textbf{Electrochemical Properties}}} \\
Ionic conductivity & Value, unit, measurement temperature \\
Coulombic efficiency & Value, unit, conditions (initial / stable / end cycles) \\
Cell operating voltage range & Range, unit, reference electrode \\
Electrolyte stability window & Range, unit, reference electrode \\
Cycling stability & Descriptive \\
Specific capacity & Value, unit \\
Energy density & Value, unit \\
Capacity retention & Descriptive \\
Overpotential & Value, unit, reference electrode (cathodic / anodic / total) \\
\midrule
\multicolumn{2}{l}{\textit{\textbf{Physicochemical Properties}}} \\
pH & Value \\
Viscosity & Value, unit, measurement temperature \\
Cell operating temperature range & Descriptive \\
\midrule
\multicolumn{2}{l}{\textit{\textbf{Provenance}}} \\
Source section & Section heading, figure, or table number \\
Source Modality & text\_only, image\_only, table\_only  \\
\bottomrule
\newline
\end{tabularx}
\caption{\label{tab:property-schema}
Target properties and captured fields in the extraction schema. Properties are grouped into composition, cell configuration, electrochemical, physicochemical, and provenance categories. Every field is nullable to avoid forcing hallucinated values when a property is not reported in the source document.
}
\end{table*}


The final list, shown in Table~\ref{tab:property-schema}, covers the core physicochemical and electrochemical descriptors needed to characterise an electrolyte--from ionic conductivity and Coulombic efficiency to stability windows, overpotentials, and capacity metrics--alongside a structured breakdown of the electrolyte composition (salts, solvents, additives) and cell information (cathode, anode). Each property is captured along with its associated metadata--value, unit, measurement conditions, and reference electrode where relevant--to preserve the context needed for downstream normalisation and comparison across papers.

Rather than leaving the LLM to produce free-form JSON, we encode this property list as an explicit JSON schema--one object per electrolyte, with nested fields for each property and its associated metadata. Every field is nullable to avoid forcing the model to hallucinate values when a property is not reported. The schema also includes a \textit{source\_section} field for every extracted entry, recording the section heading, figure number, or table number from which each value was drawn--directly addressing our grounding desideratum and enabling downstream validation and manual spot-checks.

This schema-driven design also gives the pipeline its generalizability: the property list is defined declaratively in the schema rather than hard-coded into extraction logic, meaning properties can be added, removed, or restructured without modifying the rest of the pipeline. For this paper, we target battery electrolytes and \text{$CO_2$} electro-reduction systems, but the same architecture extends naturally to adjacent electrochemical subdomains--such as fuel cells, supercapacitors, or electrochemical sensors -- by swapping in a schema tailored to that domain's properties and conventions. The document parser, summarizer, and extraction call remain unchanged.

\subsubsection{Making the API Call}
\label{subsec:callAPI}
Given the large context lengths supported by current LLMs, the entire document--structured text from \texttt{document parser} and image/table summaries from \texttt{image summarizer}---can be passed to the model in a single API call, rather than chunking the paper or processing each section in isolation. This lets the LLM reason over the full paper at once, which is essential when an electrolyte and its properties are discussed over several sections of the paper.

For each paper, we assemble the payload and pass it to the LLM along with a two-part prompt--a system prompt that instructs the model with relevant domain context (electrochem terminology, role of reference electrodes, etc.), and a user prompt that specifies the extraction task and target fields. The schemas per usecase are passed as a structured output constraint so the model is required to return a valid json conforming to the schema. 

The pipeline is designed to be model-agnostic--the same payload, schema, and prompts can be routed to any LLM that supports structured JSON outputs. We benchmark multiple models (both on closed and open-source) on this step and compare their extraction accuracy, cost and latency.

\subsection{Component E: ElectroChemDB}
\label{subsec:electroChemDB}

The LLM extraction stage produces one structured JSON object per electrolyte per paper--rich in content, but not yet ML-ready. Electrolyte and solvent names appear as free-form natural-language strings, property values reported in whichever units the original paper happened to use, and the same chemical entity can appear under multiple surface forms across different papers. The end goal is to produce a standardized, queryable data that can be consumed directly by ML pipelines--for tasks such as electrolyte property prediction, synthesis route determination--which means these inconsistencies need to be resolved before storage. To achieve this, we perform three post-processing steps: (i) chemical entity standardisation, in which molecular names are mapped to canonical SMILES representations; (ii) unit normalisation, in which extracted property values are converted to a consistent set of canonical units; and (iii) failure handling, which combines domain-specialised LLMs and expert review to resolve entities or measurements that cannot be processed automatically. The resulting records are then consolidated into ElectroChemDB, a structured and machine-learning-ready repository of electrolyte compositions and properties.

\subsubsection{Chemical Entity Standardization}
\label{subsubsec:stdz}
\begin{figure}
    \centering
    \includegraphics[width=1\linewidth]{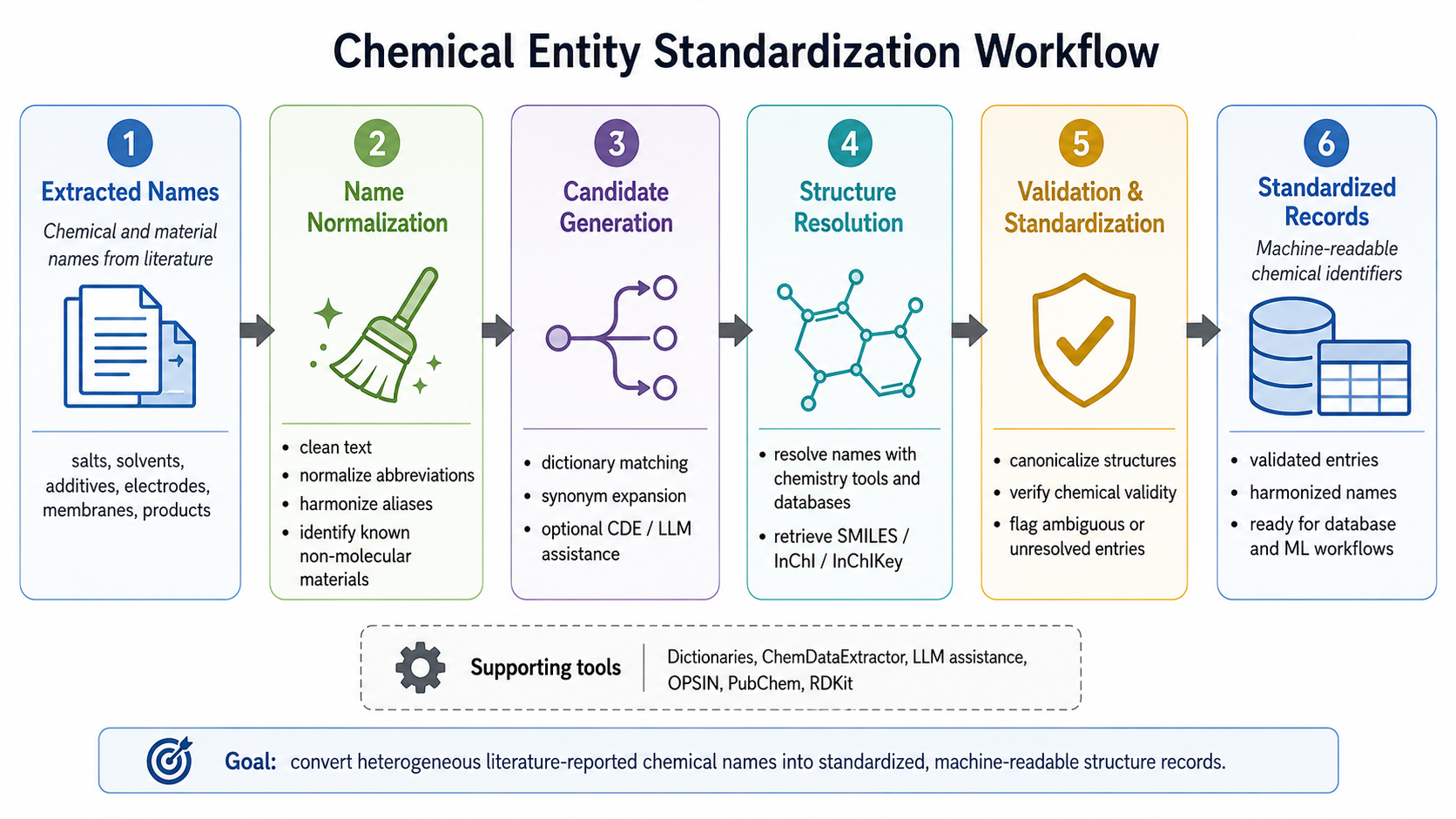}
    \caption{Chemical entity standardization workflow. Extracted literature-reported chemical and material names are normalized, expanded through dictionary and synonym-based candidate generation, resolved to molecular structures using external chemistry resources, and validated to produce canonical machine-readable identifiers for downstream database construction and machine-learning workflows.}
    \label{fig:data_standardization}
\end{figure}

For every extracted electrolyte component---including salts, solvents, and additives---we resolve the literature-reported natural-language name to a standardized molecular representation, as summarized in Fig.~\ref{fig:data_standardization}. The workflow first normalizes raw names by cleaning text, harmonizing aliases and abbreviations, and identifying known non-molecular materials that cannot be assigned a single small-molecule structure. Candidate names are then generated using lookup dictionaries, synonym expansion, ChemDataExtractor~\cite{ChemDataExtractor}, and, when needed, LLM-assisted fallback resolution. These candidates are queried against external chemistry resources, including OPSIN~\cite{opsin} for IUPAC-style names and PubChem~\cite{pubchem} for database-backed synonym and structure lookup. RDKit~\cite{Bento2020} is then used to canonicalise and validate the resulting SMILES strings~\cite{SMILES}, with unresolved or ambiguous entities flagged for review.

Standardizing chemical entities to canonical SMILES provides a single machine-readable identifier for each molecule, enabling deduplication across papers when different surface names refer to the same compound. The resulting standardized records also make ElectroChemDB directly compatible with downstream cheminformatics representations, including graph neural network inputs~\cite{GNN}, molecular embeddings, and Morgan fingerprints~\cite{roger_han}, supporting machine-learning workflows built on curated electrolyte compositions.

\subsubsection{Unit Normalisation}
\label{subsubsec:nrmz}
Extracted property values are normalised to a canonical unit per property, defined in consultation with domain experts to match the conventions most commonly used in electrochemical modelling and materials science. Normalisation is handled by a rule-based conversion layer that parses the extracted unit string and maps it to the canonical form, with robust handling of formatting variations. We also remove outliers based on the specified ranges for each property in accordance with the previous literature.

\subsubsection{Electrolyte Composition Normalization}
\label{app:composition_normalization}

Electrolyte compositions are reported in the literature using heterogeneous conventions: solvent blends appear as volume, mass, or molar ratios, while salt and additive concentrations are expressed as molarity, molality, weight or mole percentages, and other concentration units. To establish a consistent quantitative representation for downstream machine-learning applications, we apply a deterministic composition-normalization step downstream of unit normalization that converts extracted electrolyte formulations into a common molar-concentration basis (mol~L$^{-1}$). For each record, the solvent-ratio basis is identified from keywords in the reported composition, with volume ratios assumed when unspecified. Solvent ratios are converted into volume fractions using tabulated densities and molar masses where required. A nominal one-liter solvent-blend basis is then used to calculate the mass and molar amount of each solvent. Salt and additive concentrations are subsequently converted to the same basis using unit-specific relationships, while reported molarities are retained unchanged, neglecting the salt's contribution to the solution volume. Ambiguous units, missing physicochemical information, and unsuccessful conversions are recorded through component-specific status indicators, preserving the original information without introducing unsupported values. Applied to 41,412 extracted records, the procedure normalized solvent compositions in 24,847 records; the remainder failed almost entirely because the source publication reports no solvent composition or reports it incompletely. It converted 24,901 salt concentrations and 3,430 additive concentrations to the molar basis. Overall, 17,091 records (41.3\%) normalize fully across every reported component. This standardized representation establishes a consistent set of composition-dependent features for subsequent machine-learning models.

\subsubsection{Failure Handling}
\label{subsubsec:fails}
Not every extracted entity can be resolved cleanly--some molecule names are ambiguous, misspelled, or refer to proprietary or uncommon compounds--which are not listed by the chemistry tools we use for the resolution, unit strings outside the defined rule set. In these cases, we employ a general-purpose
open-weight language model (GPT-OSS-20B)~\cite{openai2025gptoss120bgptoss20bmodel} as a fallback to assist
with chemical name resolution, including name conversions
(IUPAC names to SMILES representations or molecular formulas).
Entries that remain unresolved after this step are manually
reviewed and resolved by domain experts.

\subsubsection{Storage}
\label{subsec:storage}
Standardised entries are stored in MongoDB as the back-end for ElectroChemDB. MongoDB's flexible document model accommodates schema evolution as new properties are added, while JSON Schema validation (using the schema from~\ref{subsubsec:schemas}) enforces data integrity. ElectroChemDB can be queried through MongoDB's native interface and exported as flat tabular snapshots (CSV, Parquet) for direct consumption by downstream ML pipelines.





\section{Results}
\label{app:results}

\begin{table}[ht]
\centering
\footnotesize
\setlength{\tabcolsep}{6pt}
\begin{tabularx}{\linewidth}{l X c}
\toprule
\textbf{Category} & \textbf{Property} & \textbf{F1} \\
\midrule

\textit{Composition}
& Total electrolyte composition & 94.37 \\
& Salt(s) & 98.61 \\
& Solvent(s) & 99.55 \\
& Additive(s) & 93.28 \\

\midrule
\textit{Cell Information}
& Cathode & 98.47 \\
& Anode & 98.47 \\

\midrule
\textit{Electrochemical Properties}
& Ionic conductivity & 97.66 \\
& Coulombic efficiency & 96.08 \\
& Cell operating voltage range & 98.66 \\
& Electrolyte stability window & 96.74 \\
& Specific capacity & 98.01 \\
& Energy density & 100.00 \\
& Capacity retention & 97.81 \\
& Overpotential & 90.42 \\

\midrule
\textit{Physicochemical Properties}
& pH & 90.91 \\
& Viscosity & 97.55 \\
& Cell operating temperature range & 98.17 \\

\midrule
\textbf{Overall}
& \textbf{Micro-average} & \textbf{97.15} \\

\bottomrule
\newline
\end{tabularx}
\caption{Property-wise extraction performance (F1 score) of ElectroChemLLM on the battery-electrolyte benchmark. Properties are grouped into composition, cell information, electrochemical, and physicochemical categories.}
\label{tab:property-f1}
\end{table}

\subsection{Evaluation metrics.}
\label{app:eval_metrics}
We report standard precision~\ref{eq:precision}, recall~\ref{eq:recall}, and F-1 score~\ref{eq:f1_score}, computed by comparing each extracted JSON entry against its manually annotated counterpart. A predicted value is counted as a true positive if it matches the ground-truth value after standardisation - SMILES canonicalisation~\ref{subsubsec:stdz} for chemical entities, and unit normalisation~\ref{subsubsec:nrmz} for property values. For categorical fields (e.g., reference electrode, cathode, anode), matching is exact after canonicalisation.

\begin{equation}
    \text{Precision} = \frac{\text{TP}}{\text{TP} + \text{FP}}
    \label{eq:precision}
\end{equation}

\begin{equation}
    \text{Recall} = \frac{\text{TP}}{\text{TP} + \text{FN}}
    \label{eq:recall}
\end{equation}

\begin{equation}
    F_1\text{-score} = 2 \times \frac{\text{Precision} \times \text{Recall}}{\text{Precision} + \text{Recall}} = \frac{2\text{TP}}{2\text{TP} + \text{FP} + \text{FN}}
    \label{eq:f1_score}
\end{equation}

\begin{equation}
    \text{MAE} = \frac{1}{n} \sum_{i=1}^{n} |y_i - \hat{y}_i|
    \label{eq:mae}
\end{equation}

\begin{equation}
    R^2 = 1 - \frac{\sum_{i=1}^{n} (y_i - \hat{y}_i)^2}{\sum_{i=1}^{n} (y_i - \bar{y})^2}
    \label{eq:r2_score}
\end{equation}

where the mean of the actual values $\bar{y}$ is defined as:

\begin{equation}
    \bar{y} = \frac{1}{n} \sum_{i=1}^{n} y_i
\end{equation}
\newline
We also compute F-1 at the individual property level, which surfaces which properties the pipeline extracts reliably and which remain difficult (reported in Section~\ref{tab:property-f1}).

\subsection{Cost Analysis.}
\begin{figure}
    \centering
    \includegraphics[width=0.8\linewidth]{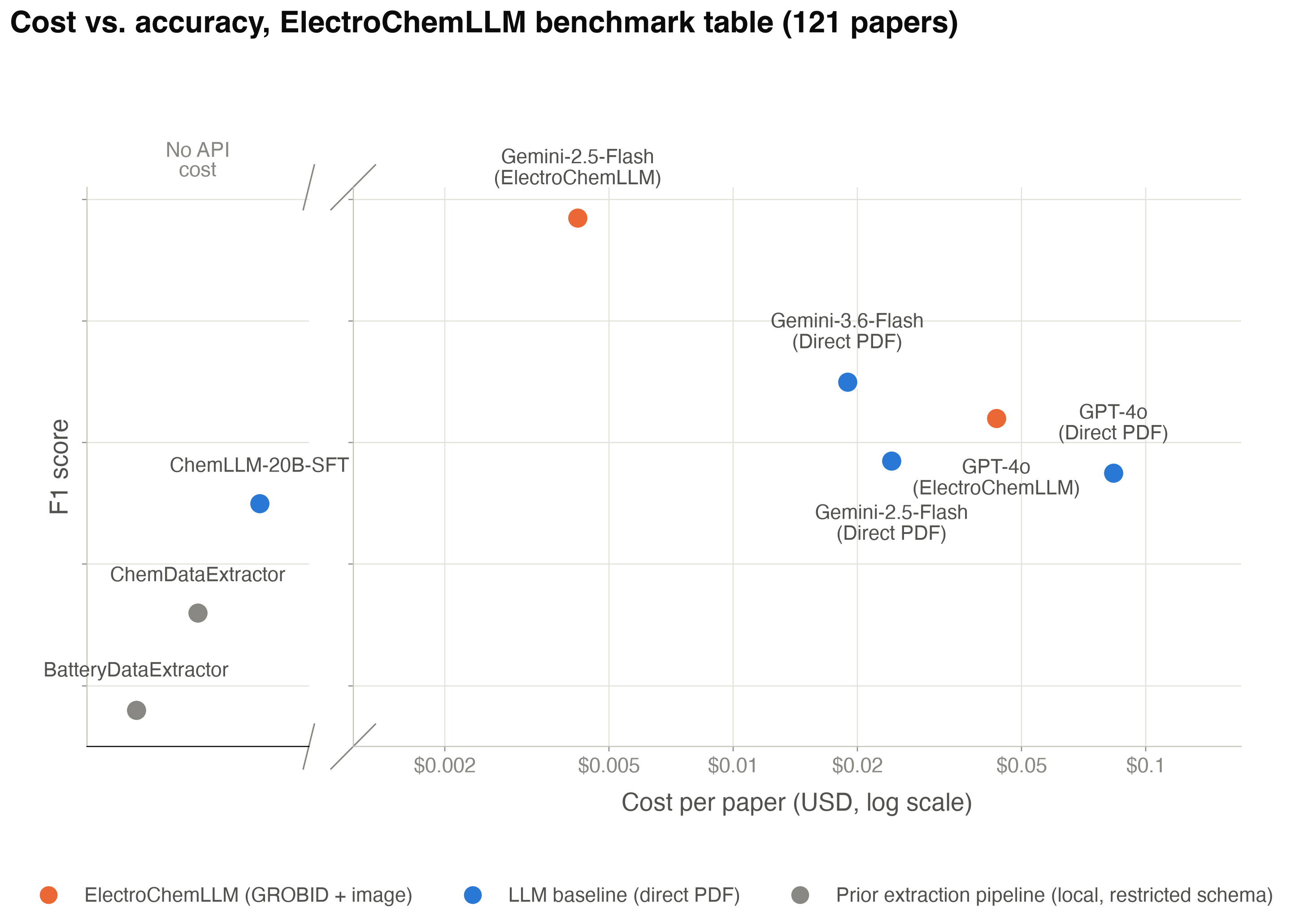}
    \caption{Extraction cost vs. accuracy across configurations on the 121-paper benchmark (Table~\ref{tab:main_results}). Points are colored by pipeline type: prior extraction pipelines (gray, local tools with no API cost), direct-PDF LLM baselines (blue), and \texttt{ElectroChemLLM} (orange). The left panel shows configurations with no API cost on a separate axis; the right panel shows per-paper API cost on a log scale. \texttt{Gemini-2.5-Flash} under \texttt{ElectroChemLLM} achieves the highest F1 (0.97) at the lowest cost among API-based methods ($\sim$\$0.004/paper), dominating all other configurations on both axes.}
    \label{fig:cost_vs_f1}
\end{figure}

Figure~\ref{fig:cost_vs_f1} situates these accuracy gains against their cost. \texttt{ElectroChemLLM} with \texttt{Gemini-2.5-Flash} is Pareto-optimal among the API-based configurations: it reaches the highest F1 (0.97) while costing roughly 4- to 20-fold less per paper than the direct-PDF baselines ($\sim$0.004 vs.\ $0.019-$0.083). The document-processing and visual-summarization components of \texttt{ElectroChemLLM} do not benefit every backbone equally -- \texttt{GPT-4o} sees only a modest F1 gain over its direct-PDF counterpart (0.64 vs.\ 0.55) at roughly half the cost, while \texttt{Gemini-2.5-Flash} gains sharply on both axes (0.97 vs.\ 0.57, at roughly one-sixth the cost of its own direct-PDF variant). This suggests the structured-input pipeline is necessary but not sufficient for accuracy gains; it interacts strongly with the underlying model's ability to exploit the pre-extracted text and visual summaries. Prior rule-based pipelines (\texttt{BatteryDataExtractor}, \texttt{ChemDataExtractor}) and the locally-run \texttt{ChemLLM-20B-SFT} incur no per-paper API cost but trail substantially on accuracy (F1 $\leq$ 0.50), underscoring the trade-off a practitioner faces between zero marginal cost and extraction quality at scale.

\subsection{Component-Wise Performance (text extraction, figure localisation, summarisation quality)}
\label{subsec:comp_perf}
Before evaluating end-to-end extraction quality, we validate that each pipeline component performs its intended function reliably. We evaluate three components individually: text extraction~\ref{subsubsec:text_extraction}, figure localisation~\ref{subsubsec:fig_loc_eval}, and image summarisation~\ref{subsubsec:img_summ_eval}.

\subsubsection{Text extraction}
\label{subsubsec:text_extraction}
To assess the performance of S2ORC-Grobid, we conduct a qualitative human review of 50 papers. Our findings suggest that S2ORC-Grobid consistently recovered the textual backbone of each article, including bibliographic metadata, abstracts, section headings, inline citations, and the majority of the narrative body text. This aligns with the original S2ORC evaluation, which reported an F1 score of approximately 0.89 for inline-citation detection and bibliography parsing \cite{lo-wang-2020-s2orc}. Localised challenges were observed for notation- and layout-sensitive content, such as chemical subscripts, mathematical symbols, equations, and occasional mixing of figure captions with body text. These cases do not limit its role for our usecase: S2ORC-Grobid provides scalable, section-aware document context, while tables, figures, and other visually structured content are handled by the complementary TableFormer and YOLO–vision summarisation components.

\subsubsection{Figure Localisation}
\label{subsubsec:fig_loc_eval}

We compare the fine-tuned YOLO sub-panel detector against Docling's baseline layout model on a randomly sampled set of 60 figures drawn from the benchmark corpus. Domain experts annotate each cropped output along two dimensions - image quality (1-3: blurry / readable / crisp) and image complexity (1-3: simple / moderate / dense) - characterising the inputs that feed into the downstream summariser. Mean scores are reported in the following table~\ref{tab:b3-eval}, with rubric details in Appendix.

\begin{table*}[t]
\centering
\small
\setlength{\tabcolsep}{6pt}
\renewcommand{\arraystretch}{1.2}
\begin{tabular}{lccc}
\toprule
\textbf{Detector} & \textbf{Accuracy}  & \textbf{Image Quality} & \textbf{Image Complexity} \\
\midrule
Fine-tuned YOLO & 91\% & 1.92 & 1.93 \\
\bottomrule
\newline
\end{tabular}
\caption{\label{tab:b3-eval}
Cropped sub-panel quality and complexity scores (Component B.3) on 60 manually annotated figures. Each dimension is rated on a 1-3 scale by domain experts; see Appendix for the full rubric.
}
\end{table*}

\subsubsection{Image Summarisation}
\label{subsubsec:img_summ_eval}
Using the same 60-figure sample, each figure-summary pair is scored by domain experts along three dimensions -- accuracy, faithfulness, and coverage -- on a 1-3 scale. The sampled figures are predominantly graph-type figures (CV plots, polarisation curves, capacity-vs-cycle curves, etc.), reflecting the dominant figure type in electrochemistry literature where quantitative property values are reported. Mean scores across the three dimensions are reported in the table~\ref{tab:c-eval}, the full rubric is in Appendix.

\begin{table*}[t]
\centering
\small
\setlength{\tabcolsep}{8pt}
\renewcommand{\arraystretch}{1.2}
\begin{tabular}{lc}
\toprule
\textbf{Dimension} & \textbf{Mean Score} \\
\midrule
Accuracy      & 2.4		\\
Faithfulness  & 2.7 \\
Coverage      & 1.8 \\
\midrule
\textbf{Overall} & \textbf{2.28} \\
\bottomrule
\newline
\end{tabular}
\caption{\label{tab:c-eval}
Image summarisation quality~\ref{subsec:image_summarizer} on 60 manually annotated graph-type figure-summary pairs. Each dimension is rated on a 1-3 scale (higher is better); see Appendix for the full rubric.
}
\end{table*}

\subsection{LLM-Judge Evaluation}
\label{app:llm_judge}

We evaluate extracted records against the human-annotated ground truth using \texttt{Gemini-3.5-Flash} as an LLM judge. The judge is run at a temperature of 0 and constrained to return structured JSON. Evaluation is performed independently for each paper through separate recall and precision passes, avoiding brittle one-to-one alignment between electrolyte records.

\paragraph{Claim Construction.}
Each ground-truth and predicted electrolyte entry is flattened into a set of non-empty scalar claims. A claim consists of a schema field and its associated value, such as a salt name, solvent ratio, ionic-conductivity value, or measurement condition. Null, empty, and unavailable values are excluded. List-valued fields, including salts, solvents, additives, and overpotentials, are treated as unordered collections.

\paragraph{Recall Pass.}
Each ground-truth electrolyte entry is compared with all predicted entries from the same paper. For every non-empty ground-truth claim, the judge determines whether an equivalent claim is recovered in the predictions. A claim is counted as recovered only when it is associated with the same electrolyte formulation.

\paragraph{Precision Pass.}
Each predicted electrolyte entry is compared with all ground-truth entries from the same paper. The judge determines whether each predicted claim is supported by the annotations. Unsupported claims and values associated with a different formulation, measurement condition, cell, voltage, or cycle count are counted as false positives.

\paragraph{Matching Criteria.}
The judge determines electrolyte identity using salts, solvents, additives, concentrations, units, and overall composition. Harmless formatting differences, such as \(1\) versus \(1.0\) or \(\mathrm{mA/cm^2}\) versus \(\mathrm{mA\,cm^{-2}}\), are accepted. Unambiguous chemical aliases and equivalent composition ratios are also accepted. However, speculative unit conversions and numerically similar but scientifically non-equivalent values are rejected. All comparisons are restricted to records from the same paper.

\paragraph{Metric Calculation.}
Let \(N_{\mathrm{GT}}\) denote the number of non-empty ground-truth claims, \(N_{\mathrm{covered}}\) the number recovered by an extraction system, \(N_{\mathrm{pred}}\) the number of non-empty predicted claims, and \(N_{\mathrm{supported}}\) the number supported by the ground truth. Recall and precision are calculated as

\begin{equation}
\mathrm{Recall}
=
\frac{N_{\mathrm{covered}}}{N_{\mathrm{GT}}},
\qquad
\mathrm{Precision}
=
\frac{N_{\mathrm{supported}}}{N_{\mathrm{pred}}}.
\end{equation}

The corresponding \(F_1\) score is

\begin{equation}
F_1
=
2
\frac{\mathrm{Precision}\times\mathrm{Recall}}
{\mathrm{Precision}+\mathrm{Recall}}.
\end{equation}

Metrics are aggregated by schema field to obtain property-level performance. The overall micro-average is calculated by pooling claim counts across fields before computing precision, recall, and \(F_1\). If a paper contains no predicted entries, its ground-truth claims are counted as false negatives. Conversely, if predicted entries have no corresponding ground-truth entries, their claims are counted as false positives.

\paragraph{Strict and Conditional Evaluation.}
The primary evaluation includes all non-empty ground-truth claims within each system's declared extraction scope. Properties that a system fails to extract remain in the recall denominator. We additionally report a conditional evaluation restricted to property--paper pairs for which the system produces at least one prediction. This secondary measure characterises performance when a system attempts an extraction, but does not measure complete retrieval coverage and is therefore reported separately from the primary results.

\subsection{Pipeline-Specific Evaluation Scopes}
\label{app:pipeline_scopes}


Several existing materials- and battery-domain models were considered as potential baselines because their extraction objectives partially overlap with the electrolyte-composition and property schema evaluated in this study. \texttt{T2BR}~\cite{T2BR2025} was considered particularly relevant because it extracts battery cell-assembly information, including electrolyte salts, solvents, and amounts; however, the trained cell-assembly checkpoint and tokenizer resources required to reproduce the published model were not available in the public repository, so it was not quantitatively evaluated. \texttt{BatteryDataExtractor}~\cite{huang2022batterydataextractor} and
\texttt{ChemDataExtractor}~\cite{ChemDataExtractor,ChemDataExtractor2}
support extraction of selected battery or materials properties, support extraction of selected battery or materials properties, but their outputs and extraction scopes do not directly match the complete structured electrolyte composition--property schema used here. \texttt{ChemLLM}~\cite{zhang2024chemllm} provides chemistry-focused LLM-based extraction, but primarily operates on extracted text and does not provide an equivalent battery-specific pipeline incorporating tables and figures. \texttt{BatteryBERT}~\cite{batterybert} supports battery-domain classification, question answering, and entity identification, but does not directly generate complete structured composition--property records. \texttt{LLaMP}~\cite{chiang2024llamp} primarily targets materials knowledge retrieval and agent-based reasoning, while \texttt{MatSciRE}~\cite{MatSciRE2024} focuses on material--property relation extraction and could not be directly evaluated because trained checkpoints were not publicly available. We also examined the \texttt{PromptDataExtraction}
polymer-property extraction pipeline~\cite{PromptDataExtraction},
which uses \texttt{MaterialsBERT}~\cite{MaterialsBERT} and LLM-based approaches.
However, the released environment and supporting data were
incomplete for direct reproduction and would require
substantial battery-specific adaptation. \texttt{PropertyExtractor}~\cite{DICL}, based on \texttt{GPT/Gemini} prompting and dynamic in-context learning, can extract user-specified material properties, but would require extensive battery-specific configuration to match our full extraction schema. Finally, \texttt{LIBAC}~\cite{LIBAC2023} provides battery-domain named-entity recognition for electrolyte-related entities such as salts, solvents, concentrations, and ratios, but does not directly link these entities into complete composition--property records. These approaches were therefore treated as related domain-specific methods rather than directly equivalent quantitative benchmarks.

All systems are evaluated on the same set of 120 papers. The LLM-based systems are evaluated on the complete schema, whereas \texttt{ChemDataExtractor2} and \texttt{BatteryDataExtractor} are evaluated only on the properties they were designed or configured to extract. \texttt{ChemDataExtractor2} is evaluated using ten custom electrolyte-property models, while \texttt{BatteryDataExtractor} is evaluated using its five default property extractors.

Free-text subfields are excluded from the rule-based evaluation because these systems do not generate descriptive text. The excluded fields comprise measurement-condition descriptions, reference-electrode descriptions, and total-electrolyte-composition summaries. Table~\ref{tab:pipeline_scopes} reports the resulting ground-truth claim counts.

\begin{table*}[t]
\centering
\footnotesize
\renewcommand{\arraystretch}{1.15}
\setlength{\tabcolsep}{6pt}
\begin{tabularx}{\textwidth}{Xrrr}
\toprule
\textbf{Property}
& \textbf{LLMs}
& \textbf{CDE2}
& \textbf{BDE} \\
\midrule
Electrolyte composition
& \num{4035} & \num{3518} & -- \\
Coulombic efficiency
& \num{1227} & \num{793} & \num{793} \\
Cell information
& \num{929} & -- & -- \\
Cell operating voltage range
& \num{688} & -- & \num{508} \\
Ionic conductivity
& \num{602} & \num{602} & \num{602} \\
Electrolyte stability window
& \num{472} & \num{327} & -- \\
Specific capacity
& \num{465} & \num{465} & \num{465} \\
Overpotential
& \num{290} & \num{218} & -- \\
Cycling stability
& \num{248} & -- & -- \\
Cell operating temperature range
& \num{242} & \num{242} & -- \\
Capacity retention
& \num{235} & \num{235} & -- \\
Viscosity
& \num{187} & \num{187} & -- \\
Energy density
& \num{56} & \num{56} & \num{56} \\
\midrule
\textbf{Total claims}
& \textbf{\num{9676}}
& \textbf{\num{6643}}
& \textbf{\num{2424}} \\
\textbf{Properties with ground-truth claims}
& \textbf{13}
& \textbf{10}
& \textbf{5} \\
\bottomrule
\end{tabularx}

\caption{Ground-truth claims included in each system's evaluation scope. Dashes indicate properties outside a system's extraction scope. The schema contains a pH field, but it is omitted because the benchmark contains no non-null pH claims.}
\label{tab:pipeline_scopes}
\end{table*}

Although several properties are shared across systems, their claim counts may differ because prose-based subfields are retained for the LLMs and excluded for the rule-based systems. For example, measurement-condition claims are included in the LLM evaluation of Coulombic efficiency, while reference-electrode claims are included in the LLM evaluation of operating-voltage ranges. The complete judge prompt, structured-output schema, model identifier, and field mappings are provided in the supplementary materials.

\end{appendices}



\end{document}